\documentclass{article}

\PassOptionsToPackage{numbers, compress}{natbib}
 \usepackage[main, final]{neurips_2026}

\usepackage[utf8]{inputenc} % allow utf-8 input
\usepackage[T1]{fontenc}    % use 8-bit T1 fonts
\usepackage{hyperref}       % hyperlinks
\usepackage{url}            % simple URL typesetting
\usepackage{booktabs}       % professional-quality tables
\usepackage{amsfonts}       % blackboard math symbols
\usepackage{nicefrac}       % compact symbols for 1/2, etc.
\usepackage{microtype}      % microtypography
\usepackage{xcolor}         % colors
\usepackage[accsupp]{axessibility}
\usepackage{multirow}
\usepackage{graphicx}
\usepackage{caption}
\usepackage[ruled,vlined]{algorithm2e}
\title{HIFC-IQA: Train-Free Cross-Domain Image Quality Assessment via Dual-Process Cognition}

\author{%
  Yu Li\textsuperscript{1,2}\thanks{Equal contribution.}\thanks{
    This work was conducted during internship in Taobao.
    Email: \texttt{lyhit2001@gmail.com}.}
  \quad
  Zhengran Shen\textsuperscript{1}\footnotemark[1]
  \quad
  Puchao Zhou\textsuperscript{1}\footnotemark[1]
  \quad
  Yukang Ding\textsuperscript{2}
  \quad
  Yachun Mi\textsuperscript{1}
  \\
  \textbf{Boyuan Yang}\textsuperscript{1}
  \quad
  \textbf{Xi Zhai}\textsuperscript{1}
  \quad
  \textbf{Shaohui Liu}\textsuperscript{1}\thanks{Corresponding author: \texttt{shliu@hit.edu.cn}}
  \\[8pt]
  \textsuperscript{1}Harbin Institute of Technology
  \quad
  \textsuperscript{2}Taobao \& Tmall Group
}

\begin{document}

\maketitle

\begin{abstract}
  % The discrepancy between controlled synthetic distortions and complex real-world degradations poses a significant domain shift challenge for No-Reference Image Quality Assessment (NR-IQA). While Unsupervised Domain Adaptation (UDA) methods aim to bridge this gap, they heavily suffer from retraining overhead and error propagation during iterative pseudo-labeling. Inspired by the dual-process theory of human cognition, we propose Hierarchical Intuitive and Fuzzy Consensus (HIFC), a completely train-free cross-domain IQA framework. HIFC employs an asymmetric perceptual decomposition to disentangle semantic and distortion features. It fuses two complementary signals from dual branches: a memory consensus branch that retrieves a stable Fuzzy Quality Centroid from a synthetic reference gallery, and an intuitive branch that captures aesthetic polarization via zero-shot vision-language alignment. Extensive experiments demonstrate that, without any parameter updates or target-domain adaptation, HIFC achieves highly competitive results against state-of-the-art UDA methods. Most notably, it exhibits exceptional robustness and generalization in the challenging synthetic-to-authentic cross-domain setting. The code will be available upon acceptance.
  The gap between controlled synthetic distortions and complex real-world degradations introduces a critical domain shift challenge for No-Reference Image Quality Assessment. Although Unsupervised Domain Adaptation (UDA) approaches attempt to address this issue, they remain constrained by substantial retraining overhead and cumulative error propagation inherent in iterative pseudo-labeling. Inspired by the dual-process theory of human cognition for remarkable generalization ability, we propose \textbf{H}ierarchical \textbf{I}ntuitive and \textbf{F}uzzy \textbf{C}onsensus (\textbf{HIFC}), a completely train-free \textbf{Cross-Domain IQA} framework. HIFC employs asymmetric perceptual decomposition to disentangle semantic content from distortion-sensitive representations. The final quality score is fused through the synergy of two complementary branches: a memory consensus branch that retrieves a stable Fuzzy Quality Centroid from a synthetic reference gallery, and an intuitive branch that captures aesthetic polarization via zero-shot vision-language alignment. 
  % Extensive experiments demonstrate that HIFC achieves highly competitive performance against state-of-the-art UDA methods across multiple benchmarks.
 Extensive experiments show that HIFC achieves state-of-the-art performance across multiple benchmarks, attaining the highest average SRCC and PLCC among competing UDA methods.
  % Most notably, HIFC exhibits superior robustness in the challenging synthetic-to-authentic cross-domain scenario. 
  Most notably, in the challenging synthetic-to-authentic cross-domain setting, HIFC achieves a substantial performance margin over competing methods, demonstrating remarkable robustness and generalization.
  Core codes will be released in \url{https://github.com/Hitly3015/HIFC-IQA}.
\end{abstract}

\section{Introduction}

With the explosive growth of visual content across online platforms, images inevitably suffer from complex degradations during acquisition, compression, and transmission. Traditional Full-Reference Image Quality Assessment~\cite{ssim,zhang2011fsim,gmsd,chen2024topiq,chen2025toward} methods, which rely on pristine reference images, are impractical in real-world scenarios such as User-Generated Content (UGC). Consequently, No-Reference Image Quality Assessment (NR-IQA)~\cite{DBCNN,hyperiqa,ke2021musiq,loda,qalign,li2024adaptive,li2025few,li2025image,chen2024q} has become the dominant research paradigm.

Despite recent progress, high-performing deep NR-IQA~\cite{hyperiqa,ke2021musiq,qalign,jia2025refine} models are typically \emph{data hungry}, requiring large-scale datasets with reliable Mean Opinion Scores (MOS). To mitigate data scarcity, many works leverage synthetically generated distortion datasets for model training~\cite{DGQA,li2023freqalign,SynIQ,UCDA,styleam}. However, this introduces a fundamental challenge: controlled synthetic distortions (e.g., Gaussian blur, JPEG compression) differ substantially from the complex, content-dependent, and composite degradations observed in real-world images. This discrepancy leads to severe domain shift, and models trained on synthetic data often exhibit dramatic performance degradation when directly applied to authentic distortions~\cite{ieiqa,ofiqa}.

To bridge the \emph{synthetic-to-authentic} gap, \textbf{Cross-Domain Image Quality Assessment} (CD-IQA)~\cite{styleam,DGQA,SynIQ,graphiq} has recently attracted increasing attention. The prevailing solutions largely rely on Unsupervised Domain Adaptation (UDA)~\cite{DANN,MMD}. Some approaches attempt to align feature distributions between source and target domains via adversarial learning, while others adopt pseudo-labeling strategies~\cite{SynIQ} to generate soft supervisory signals for iterative fine-tuning on the target domain.
\label{sec:intro}
\begin{figure}[t]
    \centering
    \includegraphics[width=\linewidth]{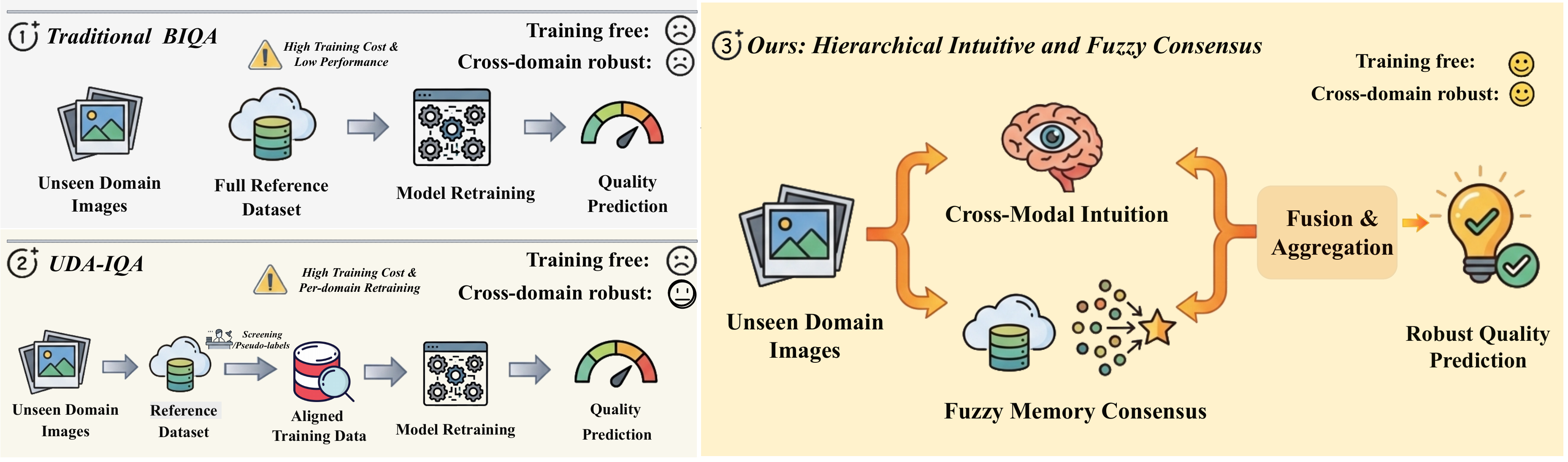}
    \caption{Comparison between previous methods and our HIFC. Unlike conventional Blind Image Quality Assessment (BIQA) and UDA-based methods requiring costly retraining or domain alignment, HIFC mimics the dual-process mechanism of the HVS by fusing zero-shot intuition with fuzzy memory consensus.
    % Motivation of the proposed HIFC framework. Unlike conventional BIQA and UDA-based methods that rely on costly retraining or target-specific distribution alignment, HIFC mimics the dual-process mechanism of the HVS by integrating zero-shot intuition with fuzzy memory consensus.
    % our framework achieves robust generalization across diverse domains without any parameter updates, offering a truly train-free solution for cross-domain IQA.
    }
    \label{fig:motivation}
\end{figure}
Although UDA-based methods have demonstrated improvements on specific benchmarks, they present notable limitations in large-scale or dynamic deployment scenarios. First, they incur substantial retraining overhead. Existing UDA frameworks are typically domain-specific: when encountering a new, unseen target domain, the model must undergo a costly adaptation cycle involving interdependent components such as data filtering or pseudo-label generation to facilitate iterative fine-tuning. This requirement essentially undermines plug-and-play applicability. 
% Second, pseudo-labeling is inherently vulnerable to noise. Given the complexity of authentic distortions, initial predictions on the target domain are often unreliable; such errors, even after filtering, are progressively amplified during iterative updates, potentially leading the model toward a sub-optimal local minimum. 
More fundamentally, this "adaptation-per-domain" paradigm deviates from the way humans assess visual quality. Inspired by dual-process theory~\cite{kahneman2011thinking} of Human Visual System(HVS), we revisit a fundamental question: \emph{Does evaluating an unseen distorted image truly require updating model parameters for every new domain?} Humans exhibit remarkable generalization ability without explicit retraining. When encountering unfamiliar distortions, perception integrates rapid intuitive judgments with comparisons to previously experienced distortion prototypes~\cite{nosofsky1986attention} stored in memory, forming a fuzzy conceptual prototype (i.e., a centroid)~\cite{rosch1975cognitive} rather than seeking exact point-to-point alignments.

Motivated by this observation, as shown in Fig.~\ref{fig:motivation}, we propose a fully train-free cross-domain IQA framework termed HIFC (Hierarchical Intuitive and Fuzzy Consensus). HIFC eliminates the need for domain adaptation or target-domain fine-tuning. Specifically, we first perform an asymmetric perceptual decomposition to disentangle semantic and distortion-related representations. 

In the \emph{memory consensus} branch, distortion features are leveraged to perform hierarchical retrieval within a pre-constructed synthetic reference gallery. By aligning topological consistency and aggregating retrieved prototypes, we compute a \emph{Fuzzy Quality Centroid (FQC)} that provides robust objective calibration.

In parallel, the \emph{intuitive perception} branch exploits zero-shot alignment capabilities in a vision-language space to obtain a subjective aesthetic polarization score. The final quality prediction is obtained through a linear consensus of the memory-based calibration and intuitive assessment, forming a closed-loop evaluation mechanism that integrates experience and intuition.

Our main contributions are summarized as follows:

\begin{itemize}

\item We propose an efficient Train-free paradigm for Cross-Domain IQA. The proposed method eliminates the need for complex parameter fine-tuning or pseudo-label generation for a specific target domain, and provides a more practical option for quality assessment in dynamic environments.

\item Inspired by the dual-process mechanism of the human visual system, we propose a dual-branch framework that disentangles distortion and semantic representations. Combined with structured memory retrieval, the design mitigates domain shift and improves cross-domain image quality assessment.

\item 
% Extensive experiments across multiple cross-domain settings demonstrate that HIFC achieves consistent and competitive performance compared with existing adaptation-based approaches. Moreover, as a train-free framework, HIFC can be deployed without additional retraining or modification to target domains.
Extensive experiments across multiple cross-domain settings demonstrate that HIFC, without any parameter updates or target-domain adaptation, achieves superior performance over state-of-the-art adaptation-based methods. As a fully train-free framework, HIFC offers plug-and-play deployability across diverse domains.
% , eliminating the need for retraining or architectural modification entirely.
    
\end{itemize}

\section{Related Works}
\textbf{No Reference Image Quality Assessment:} Deep learning has significantly advanced Blind Image Quality Assessment (BIQA) by achieving superior accuracy through complex architectural innovations~\cite{DBCNN,Tres,chen2024promptiqa,loda}. Various paradigms have been explored to enhance performance: Zhu et al.~\cite{zhu2020metaiqa} and Su et al.~\cite{hyperiqa} utilized meta-learning and self-adaptive hyper-networks, respectively, to handle diverse distortions. Meanwhile, transformer-based architectures like MUSIQ~\cite{ke2021musiq} and local feature injected ViTs~\cite{loda} have been developed to capture multi-scale or fine-grained degradation. Other works have focused on representation learning, such as Re-IQA~\cite{saha2023re}, which employs unsupervised mixture-of-experts, and QCN~\cite{qcn}, which utilizes comparison transformers to improve generalization.  However, a fundamental limitation persists: these high-performing models are inherently data-hungry and training-dependent. More importantly, even methods targeting generalization still operate under a "weight-tuning" paradigm. This heavy reliance on supervised domain-specific fine-tuning motivates a shift toward a more efficient, train-free approach that mimics the human visual system's ability to assess quality through rapid intuition and prior experience without the need for constant parameter updates.\\

\noindent
\textbf{Unsupervised Domain Adaptation Image Quality Assessment:} Due to the significant domain differences between synthetic distortions and real-world distortions, models trained on synthetic data often perform poorly when facing real-world images. To bridge this gap, several studies have explored Unsupervised Domain Adaptation (UDA) techniques. Chen et al.~\cite{UCDA} proposed a curriculum-style UDA approach for video quality assessment, adapting models progressively. Lu et al.~\cite{styleam} introduced StyleAM, aligning source and target domains in the feature style space. Li et al.~\cite{li2023freqalign} developed FreqAlign to excavate transferability from a frequency perspective, while ~\cite{DGQA} proposed DGQA, a distortion-guided UDA framework that matches data distributions via adaptive multi-domain selection. Recently, more advanced paradigms have emerged to tackle the complexities of domain shift. Meanwhile, SynDR-IQA~\cite{SynIQ} attempts to enhance the syn-to-real generalization by reshaping the distribution of synthetic data to better align with authentic environments. However, a common limitation persists: previous UDA-based methods are fundamentally rooted in an adaptation-per-domain paradigm. These requirements introduce substantial retraining overhead and limit their utility in dynamic scenarios. Motivated by the remarkable generalization capacity of the Human Visual System (HVS) in perceptual quality judgment, we propose a completely train-free framework to the challenges outlined above. 

%###Yukang adding：
% Motivated by the remarkable generalization capacity of the Human Visual System (HVS) in perceptual quality judgment, we propose a completely train-free framework-HIFC to the challenges outlined above.

\section{Methodology}
\subsection{Overview}
\begin{figure}[t]
    \centering
    \includegraphics[width=\linewidth]{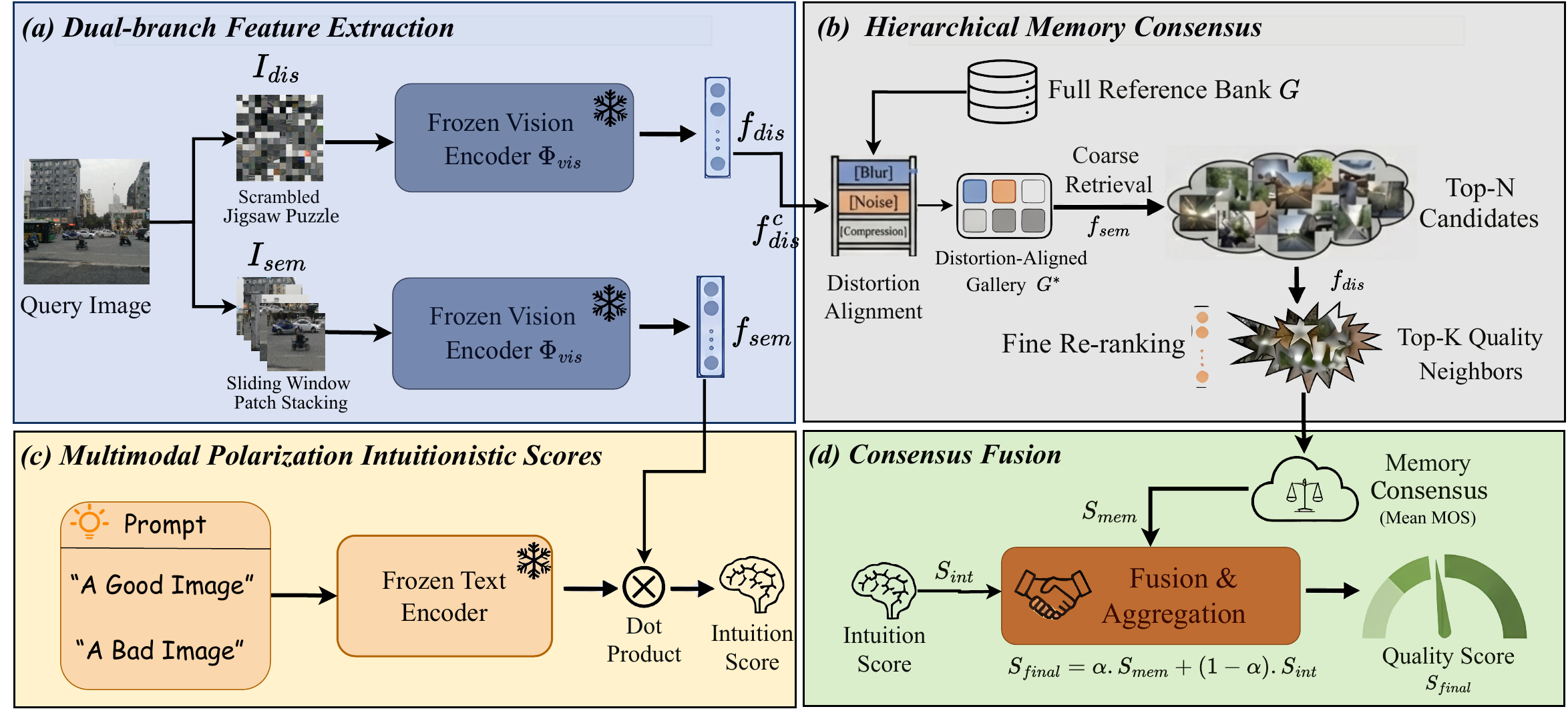}
    \caption{The illustration of the HIFC-IQA framework. The proposed architecture employs an asymmetric perceptual decomposition to decouple semantic and distortion-related features, enabling a dual-branch synergy between intuitive score and memory-based fuzzy consensus.}
    \label{fig:pipeline}
\end{figure}

The overall pipeline of the proposed HIFC (Hierarchical Intuitive and Fuzzy Consensus) framework is illustrated in Fig.~\ref{fig:pipeline}. The framework follows a paradigm of asymmetric perceptual decomposition, hierarchical memory retrieval, and dual-process consensus fusion. Its core idea is to leverage frozen multimodal priors to decouple semantic content from distortion, enabling quality prediction through the integration of memory-based calibration and prototype-based alignment.

Given an input image $I \in \mathbb{R}^{H \times W \times 3}$, two asymmetric pre-processing strategies are applied to separately emphasize semantic and distortion information. The resulting inputs are encoded by a shared frozen visual encoder $\Phi_{\text{vis}}$, producing semantic-aware feature $f_{\text{sem}}$ and distortion-aware feature $f_{\text{dis}}$.
In the memory consensus path, distortion alignment is first performed to filter the global reference gallery $G$ and construct a distortion-consistent sub-gallery $G^\ast$. A hierarchical retrieval strategy is then applied: semantic coarse filtering selects the Top-$N$ candidates in the $F_{\text{sem}}$ space, followed by distortion-aware re-ranking in the $F_{\text{dis}}$ space. The Fuzzy Quality Centroid (FQC) score $S_{\text{mem}}$ is computed by aggregating the MOS of the Top-$K$ neighbors, providing a memory-grounded quality estimate.

In parallel, the intuition path projects $f_{\text{sem}}$ into a vision-language space to compute a Binary Polarization score $S_{\text{int}}$ with respect to quality prototypes. Finally, a weighted fusion integrates the intuition and memory scores to produce the final prediction. By mimicking the dual-process mechanism of human cognition, HIFC achieves robust cross-domain performance without additional training overhead.

\subsection{Asymmetric Dual-Branch Feature Extraction}

The perceptual quality of an image is inherently characterized by the coupling of global semantic context and local distortion artifacts. To comprehensively capture quality-related information while disentangling these two components, we propose an asymmetric dual-branch feature extraction scheme using a shared frozen encoder $\phi$, as illustrated in Fig.~\ref{fig:pipeline}(a).

The distortion-aware branch aims to extract content-agnostic quality representations $f_{\text{dis}}$. Inspired by grid-based sampling strategies in VQA~\cite{fastvqa} for capturing local details, we employ a \emph{Stochastic Jigsaw} preprocessing method to achieve semantic deconstruction.

Given a grid size $G_{\text{size}}$, the original image $I$ is first partitioned into a set of localized patches $\{p_{i,j}\}$. Considering that synthetic distortions in large-scale databases (e.g., KADID-10k) are typically spatially homogeneous, we apply a grid-based shuffling operator $\mathcal{S}$ to disrupt the spatial topology and remove global semantic structure. The process is formulated as:
\begin{equation}
I_{\text{dis}} = \mathcal{S}\left( \{ p_{i,j} \mid p_{i,j} = \text{Crop}(I, \text{pos}_{i,j}, G_{\text{size}}) \} \right).
\end{equation}

The scrambled image $I_{\text{dis}}$ is then fed into the frozen encoder $\phi$ to extract the distortion-aware feature:
\begin{equation}
f_{\text{dis}} = \text{Norm}\big(\phi(I_{\text{dis}})\big).
\end{equation}

The semantic-aware branch focuses on capturing category information and contextual background, as human sensitivity to the same distortion varies significantly across different semantic carriers.

To preserve semantic integrity while avoiding distortions introduced by disproportionate scaling or information loss from center cropping, we adopt a  non-overlapping sliding-window stacking strategy. Let $\mathcal{W}$ denote the sliding window operator that extracts $M$ non-overlapping patches $\{w_1, w_2, \dots, w_M\}$ from the original image. The stacked semantic input $I_{\text{sem}}$ is defined as:
\begin{equation}
I_{\text{sem}} = \text{Stack}(\mathcal{W}(I)) = \text{Concat}(w_1, w_2, \dots, w_M).
\end{equation}

The semantic feature $f_{\text{sem}}$ is then obtained by feeding $I_{\text{sem}}$ into the encoder $\phi$, followed by a global average pooling (GAP) operation to aggregate contextual information:
\begin{equation}
f_{\text{sem}} = \text{GAP}\big(\phi(I_{\text{sem}})\big).
\end{equation}

Through the above asymmetric processing,  the original input image is projected into two spaces with independent physical meaning. We are able to find semantically similar samples by $f_{sem}$ and distortively similar neighbors by $f_{dis}$.

\subsection{Hierarchical Memory Consensus}
To simulate the fuzzy memory search process humans employ when encountering images from an unknown domain, we formulate this mechanism as a calibration task using a structured synthetic distortion memory bank. This process involves two stages: Distortion Alignment and Memory Consensus Retrieval.

\textbf{Distortion Alignment}: In traditional UDA-based IQA, synthetic datasets typically contain a much broader spectrum of distortion types than real-world scenarios. Aligning distortion distributions between the reference gallery and the target domain is therefore essential for cross-domain perceptual consistency. Unlike prior UDA approaches that train dedicated distortion classifiers on the reference bank, our distortion features $\mathbf{f}_{\text{dis}}$, extracted via stochastic jigsaw preprocessing, inherently encode distortion-aware characteristics.

We perform domain-level alignment by matching the centroid of the target domain, $\mathbf{f}^{c}_{\text{dis}}$, with the prototypical centroids $\mathbf{f}^{rc}_{\text{dis},i}$ of each distortion category $i$ in the reference gallery. The Top-$n_{\text{dist}}$ most relevant distortion types are selected to construct a task-aligned sub-gallery $\mathcal{G}^\ast$:
\begin{equation}
\mathcal{G}^\ast = \bigcup_{i \in \mathcal{J}^\ast} \mathcal{G}_i, 
\quad 
\mathcal{J}^\ast = \arg\text{top-}n_{\text{dist}} 
\left( 
\cos(\mathbf{f}^{c}_{\text{dis}}, \mathbf{f}^{rc}_{\text{dis},i})
\right),
\end{equation}
where $\mathcal{J}^\ast$ denotes the index set of the selected distortion categories. By leveraging the zero-shot priors of the frozen visual encoder, we obtain a domain-aligned memory space $\mathcal{G}^\ast$ without any parameter updates.

\textbf{Hierarchical Semantic-Distortion Retrieval}: Within the aligned sub-gallery $\mathcal{G}^\ast$, we adopt a hierarchical retrieval strategy consisting of semantic coarse filtering followed by distortion-based fine re-ranking. Given the semantic feature $\mathbf{f}^{Q}_{\text{sem}}$ of a query image, we first retrieve a candidate set $\mathcal{N}_{\text{sem}}$ containing the Top-$N$ semantically similar samples:
\begin{equation}
\mathcal{N}_{\text{sem}} = 
\left\{
g \in \mathcal{G}^\ast 
\;\middle|\;
\text{rank}\big(\cos(\mathbf{f}^{Q}_{\text{sem}}, \mathbf{f}^{g}_{\text{sem}})\big) \leq N
\right\}.
\end{equation}

This step ensures content consistency between the query and the references, mitigating quality biases introduced by semantic sensitivity. However, semantic clustering alone may lead to ``semantic obsession,'' where content similarity dominates quality perception. Inspired by re-ranking mechanisms in Retrieval-Augmented Generation (RAG), we conduct a second-stage ranking within $\mathcal{N}_{\text{sem}}$ using distortion features $\mathbf{f}_{\text{dis}}$. The Top-$K$ neighbors are selected as:
\begin{equation}
\text{FQC} =
\left\{
k \in \mathcal{N}_{\text{sem}}
\;\middle|\;
\text{rank}\big(\cos(\mathbf{f}^{Q}_{\text{dis}}, \mathbf{f}^{k}_{\text{dis}})\big) \leq K
\right\}.
\end{equation}

Finally, we define the Fuzzy Quality Cluster as the set of Top-$K$ retrieved anchors. The memory consensus score $S_{\text{mem}}$ is computed as the centroid of their subjective scores in label space:
\begin{equation}
S_{\text{mem}} = \frac{1}{K} \sum_{k \in \text{FQC}} s_k,
\end{equation}
where $s_k$ denotes the subjective quality score of the $k$-th reference sample. 

This hierarchical calibration process provides an objective quality benchmark grounded in the local topology of the memory manifold, effectively bridging the synthetic-to-authentic domain gap through stable topological consensus.

\subsection{Polarization Intuition and Final Fusion}

Inspired by the System 1 mechanism of the HVS~\cite{kahneman2011thinking}, humans often form an instantaneous and affective binary judgment of image quality before engaging in deliberate reasoning. In HIFC, we model this behavior as a Polarization Process, where image features gravitate toward two semantic poles "good" and "bad" within a multimodal embedding space.

We define a set of textual prompts $p_c$, where $c \in \{\text{pos}, \text{neg}\}$ corresponds to ``high-quality image'' and ``low-quality image,'' respectively. Using a frozen text encoder $\Psi_{\text{txt}}$, these prompts are projected onto the unit hypersphere as textual anchors:
\begin{equation}
\mathbf{w}_c = \text{Norm}\big(\Psi_{\text{txt}}(p_c)\big), 
\quad c \in \{\text{pos}, \text{neg}\}.
\end{equation}

Given the query semantic feature $\mathbf{f}^{Q}_{\text{sem}}$, intuitive quality estimation is formulated as perceptual alignment with these two prototypes rather than direct regression. A temperature-scaled Softmax operator is adopted to model the binary intuition probability:
\begin{equation}
S_{\text{int}} =
\frac{
\exp\left(\tau \cdot \cos(\mathbf{f}^{Q}_{\text{sem}}, \mathbf{w}_{\text{pos}})\right)
}{
\sum_{c \in \{\text{pos}, \text{neg}\}}
\exp\left(\tau \cdot \cos(\mathbf{f}^{Q}_{\text{sem}}, \mathbf{w}_c)\right)
},
\end{equation}
where $\tau$ denotes the temperature scaling factor. The score $S_{\text{int}}$ reflects the zero-shot confidence that the image gravitates toward the positive semantic pole. This polarization mechanism captures macro-aesthetic perception and complements the fine-grained degradation assessment provided by the memory path.

HIFC integrates intuition and rational memory through a linear consensus operator, simulating the cognitive progression from an initial first-glance impression to experience-based calibration. The final quality prediction $S_{\text{final}}$ is defined as:
\begin{equation}
S_{\text{final}} =
\alpha \cdot S_{\text{mem}} + (1 - \alpha) \cdot S_{\text{int}},
\end{equation}
where $\alpha$ balances the contributions of the two pathways. Through this dual-process mechanism, HIFC preserves its training-free nature while simultaneously modeling micro-level technical degradations via $S_{\text{mem}}$ and macro-level perceptual preferences via $S_{\text{int}}$.

\section{Experiments}
\subsection{Experimental Setups}
\textbf{Datasets and Protocols}: We conduct experiments on several IQA datasets: four traditional synthetic distortion datasets LIVE~\cite{live}, CSIQ~\cite{csiq}, TID2013~\cite{tid2013}, KADID-10k~\cite{kadid-10k}, four authentic distortion datasets LIVEC~\cite{livec}, KonIQ-10k~\cite{hosu2020koniq}, BID~\cite{bid}, and the dataset PIPAL~\cite{gu2020pipal} with both synthetic and algorithmic distortions. Model performance is assessed using Spearman’s Rank Correlation Coefficient(SRCC) and Pearson’s Linear Correlation Coefficient(PLCC). Both coefficients range from -1 to 1, with values closer to 1 indicating better performance.\\
\textbf{Implementation Details}: HIFC utilizes the frozen CLIP (ViT-B/32)~\cite{clip} as the unified backbone. 
For asymmetric preprocessing, the semantic branch extracts non-overlapping $224 \times 224$ windows with a stride of $224$, while the distortion branch adopts a Scrambled Jigsaw strategy with grid size $G_{\text{size}} = 16$. During hierarchical retrieval, the coarse candidate set size is fixed to $N = 1000$. Instead of using a predefined Top-$K$, we employ an adaptive thresholding strategy, where the FQC is formed by selecting samples whose similarity exceeds the mean similarity within the candidate set. The final fusion weight \(\alpha\) is set to $0.8$. More implementation details and reasons for parameter selection can be found in the Appendix Sec.~\ref{sec:add_exp_setup}.\\

\subsection{Performance Comparison of SOTA Methods}
\textbf{Performance on the Synthetic-to-Authentic Setting:} We first evaluate the generalization capability of HIFC when transferring from synthetic to authentic distortions. The results of the comparison between HIFC-IQA and state-of-the-art(SOTA) methods are summarized in Table~\ref{tab:all_results}. HIFC achieves superior performance on the majority of benchmarks, obtaining the highest average SRCC and PLCC among all competing methods. Compared to UDA-based methods (e.g., DGQA, SynDR-IQA), which require costly adaptation cycles, HIFC delivers superior accuracy through a purely train-free paradigm. This plug-and-play capability highlights the efficiency of our hierarchical retrieval mechanism in bridging the domain gap without retraining overhead. Furthermore, HIFC significantly surpasses VLM-based models like Q-Align and CLIPIQA+. This performance gap demonstrates that our success is not merely a byproduct of powerful pre-trained backbones, but a result of the effective synergy between affective intuition and fuzzy memory calibration. While standard VLMs rely on global semantic encoding, our asymmetric decoupling effectively captures fine-grained authentic degradations by simulating the dual-process mechanism of HVS. These findings validate that HIFC enables highly accurate cross-domain quality prediction without any parameter updates. Moreover, analogous to Retrieval-based frameworks, HIFC can further enhance its performance by fine-tuning specific hyperparameters such as $n_{dist}$ and $\mathcal{K}$ for target datasets. More extensive experimental results are provided in the Sec.~\ref{sec:app_qua} and Sec.~\ref{sec:app_cross} in Appendix.
\begin{table}[t]
\centering
\caption{Performance comparison on the synthetic-to-authentic setting
(KADID-10k$\rightarrow$LIVEC, KonIQ-10k, and BID).
Train indicates whether IQA-specific parameter training is required.}
\label{tab:all_results}
\resizebox{0.95\columnwidth}{!}{
\begin{tabular}{l c cc cc cc cc}
\toprule
\multirow{2}{*}{Methods}
& \multirow{2}{*}{Train}
& \multicolumn{2}{c}{LIVEC~\cite{livec}}
& \multicolumn{2}{c}{KonIQ-10k~\cite{hosu2020koniq}}
& \multicolumn{2}{c}{BID~\cite{bid}}
& \multicolumn{2}{c}{Average} \\
\cmidrule(lr){3-4}
\cmidrule(lr){5-6}
\cmidrule(lr){7-8}
\cmidrule(lr){9-10}
& & SRCC & PLCC & SRCC & PLCC & SRCC & PLCC & SRCC & PLCC \\
\midrule

RankIQA~\cite{liu2017rankiqa}
& Yes & 0.491 & 0.495 & 0.603 & 0.551 & 0.510 & 0.367 & 0.535 & 0.471 \\

DBCNN~\cite{DBCNN}
& Yes & 0.572 & 0.589 & 0.639 & 0.618 & 0.620 & 0.609 & 0.613 & 0.606 \\

HyperIQA~\cite{hyperiqa}
& Yes & 0.490 & 0.487 & 0.545 & 0.556 & 0.379 & 0.282 & 0.472 & 0.442 \\

MUSIQ~\cite{ke2021musiq}
& Yes & 0.517 & 0.524 & 0.554 & 0.573 & 0.575 & 0.600 & 0.549 & 0.566 \\

VCRNet~\cite{pan2022vcrnet}
& Yes & 0.561 & 0.548 & 0.517 & 0.525 & 0.542 & 0.545 & 0.540 & 0.540 \\

KGANet~\cite{zhou2024multitask}
& Yes & 0.575 & -- & 0.528 & -- & -- & -- & -- & -- \\

CLIPIQA+~\cite{clipiqa}
& Yes & 0.512 & 0.543 & 0.511 & 0.515 & 0.474 & 0.442 & 0.499 & 0.500 \\

Q-Align~\cite{qalign}
& Yes & 0.702 & 0.744 & 0.668 & 0.665 & 0.576 & 0.651 & 0.649 & 0.678 \\

\midrule

DANN~\cite{DANN}
& Yes & 0.499 & 0.484 & 0.638 & 0.636 & 0.586 & 0.510 & 0.574 & 0.543 \\

UCDA~\cite{UCDA}
& Yes & 0.382 & 0.358 & 0.496 & 0.501 & 0.348 & 0.391 & 0.408 & 0.417 \\

RankDA~\cite{RANKDA}
& Yes & 0.451 & 0.455 & 0.638 & 0.623 & 0.535 & 0.582 & 0.542 & 0.553 \\

StyleAM~\cite{styleam}
& Yes & 0.584 & 0.561 & 0.700 & 0.673 & 0.637 & 0.567 & 0.640 & 0.600 \\

FreqAlign~\cite{li2023freqalign}
& Yes & 0.618 & 0.588 & 0.748 & 0.721 & 0.674 & 0.708 & 0.680 & 0.673 \\

DGQA~\cite{DGQA}
& Yes & 0.696 & 0.690 & 0.681 & 0.687 & 0.770 & 0.753 & 0.716 & 0.710 \\

SynDR-IQA~\cite{SynIQ}
& Yes & 0.713 & 0.714 & 0.727 & 0.735 & \textbf{0.788} & 0.764 & 0.743 & 0.737 \\

GDCIQA~\cite{graphiq}
& Yes & 0.753 & 0.756 & 0.721 & 0.738 & 0.787 & \textbf{0.781} & 0.754 & 0.757 \\

\midrule

Ours
& \textbf{No}
& \textbf{0.778} & \textbf{0.780}
& \textbf{0.765} & \textbf{0.759}
& 0.744 & 0.749
& \textbf{0.762} & \textbf{0.763} \\

\bottomrule
\end{tabular}
}
\end{table}
\begin{table}[h]
\centering
\caption{Performance comparison of different distortions between DGQA and Ours on PIPAL~\cite{gu2020pipal}.}
\label{tab:algorithmic}
\resizebox{0.6\columnwidth}{!}{
\begin{tabular}{l cc cc}
\toprule
\multirow{2}{*}{Distortion Type} 
& \multicolumn{2}{c}{DGQA~\cite{DGQA}} 
& \multicolumn{2}{c}{Ours} \\
\cmidrule(lr){2-3} \cmidrule(lr){4-5}
& SRCC & PLCC & SRCC & PLCC \\
\midrule
Traditional SR
& 0.4897 & 0.4567 
& \textbf{0.4981} 
& \textbf{0.4939} \\

PSNR-originated SR 
& 0.5419 & 0.5404 
& \textbf{0.5627} 
& \textbf{0.5639} \\

SR with kernel mismatch 
& 0.5810 & 0.5956 
& \textbf{0.5892} 
& \textbf{0.5981} \\

GAN-based SR
& 0.1629 & 0.1353 
& \textbf{0.3020} 
& \textbf{0.3010} \\

Denoising
& 0.5393 & 0.5279 
& \textbf{0.6172} 
& \textbf{0.5912} \\

SR and Denoising Joint 
& 0.5588 & 0.5193 
& \textbf{0.5594}  
& \textbf{0.5205}  \\

\midrule
ALL 
& 0.4789 & 0.4625 
& \textbf{0.5071} 
& \textbf{0.5114} \\
\bottomrule
\end{tabular}
}
\end{table}

\textbf{Performance on the Synthetic-to-Algorithmic Setting:} We further assess the generalization ability of HIFC in the synthetic-to-algorithmic setting. Table~\ref{tab:algorithmic} compares our approach with DGQA across multiple categories of algorithmic distortions. HIFC consistently surpasses DGQA in all evaluated scenarios, demonstrating stronger robustness to complex artifacts introduced by modern image processing pipelines. The performance improvement is particularly pronounced in challenging cases. For example, on GAN-based super-resolution, HIFC achieves an approximately 14\% gain in SRCC, and nearly an 8\% improvement on Denoising tasks. These distortions often exhibit non-natural structural artifacts that are difficult for conventional UDA methods to align effectively. These results suggest that the integration of affective intuition with hierarchical memory calibration enables HIFC to better model the perceptual characteristics of algorithmically generated artifacts, confirming its robustness across diverse and previously unseen distortion manifolds.

\begin{table}[t]
\centering
\caption{Performance comparison on the synthetic-to-synthetic setting(KADID-10k$\rightarrow$LIVE, CSIQ, and TID2013).}
\label{tab:synthetic}
\resizebox{0.8\columnwidth}{!}{
\begin{tabular}{lccc|cc|cc}
\toprule
\multirow{2}{*}{Methods}&\multirow{2}{*}{Training} 
& \multicolumn{2}{c|}{LIVE} 
& \multicolumn{2}{c|}{CSIQ} 
& \multicolumn{2}{c}{TID2013} \\
& & SRCC & PLCC 
& SRCC & PLCC 
& SRCC & PLCC \\
\midrule
DGQA &$\checkmark$ 
& 0.9173 & 0.8988 
& 0.7965 & 0.8017 
& 0.7077 & 0.7220 \\
SynDR-IQA 
&$\checkmark$& \textbf{0.9258} & \textbf{0.9014} 
& 0.8069 & 0.8092
& \textbf{0.7147} & \textbf{0.7328} \\
\midrule
CLIP-IQA &$\times$& 0.8769 & 0.8703 
& 0.7664 & 0.8027 
& 0.6688 & 0.6873 \\
Ours 
&$\times$& 0.9036 & 0.8913 
& \textbf{0.8113} & \textbf{0.8359} 
& 0.7041 & 0.6794 \\
\bottomrule
\end{tabular}
}
\end{table}

\textbf{Performance on the Synthetic-to-Synthetic Setting:} We also evaluate HIFC on the synthetic-to-synthetic setting. The SRCC and PLCC results are summarized in Table~\ref{tab:synthetic}. Compared to the training-free baseline CLIP-IQA, HIFC demonstrates significant improvements across all evaluated datasets. Notably, on the CSIQ dataset, our method achieves an SRCC of 0.8113, outperforming even the fully supervised SOTA method SynDR-IQA. While maintaining highly competitive performance on LIVE and TID2013, HIFC significantly narrows the gap with training-based approaches. These results confirm that our Hierarchical dual-process mechanism effectively extracts robust and generalized quality representations without requiring any parameter fine-tuning.
\subsection{Qualitative Analysis}
% To better demonstrate the effectiveness of HIFC, we conducted some qualitative analyses to provide an intuitive proof. More qualitative experiments can be found in Sec.~\ref{sec:app_qua} of the Appendix.

% \textbf{gMAD Competition with DGQA:} To intuitively evaluate the perceptual alignment and robustness of our proposed framework, we conduct a group Maximum Differentiation (gMAD) competition~\cite{gmad} between HIFC and DGQA. Fig.~\ref{fig:gmad_dgqa} illustrates the gMAD pairs generated by the competition. In the right column, where HIFC acts as the defender and DGQA as the attacker, the paired images exhibit highly similar visual quality levels despite containing different content and degradation types. This indicates that HIFC successfully resists the attack from DGQA, maintaining stable and accurate predictions. Conversely, when the roles are reversed and DGQA acts as the defender (left column), HIFC consistently identifies the severe failure cases of DGQA which display substantial differences in quality according to human perception, pairing relatively clear, high-quality scenes with heavily distorted, blocky artifacts, yet DGQA incorrectly rates them as having the same quality. These qualitative findings strongly align with our quantitative results, suggesting that by leveraging the synergy of affective intuition and memory consensus, HIFC generalizes significantly better to complex, unseen distortions and aligns more closely with the human visual system than heavily optimized training-based models.
\begin{figure}[t]
    \centering
    \includegraphics[width=0.8\linewidth]{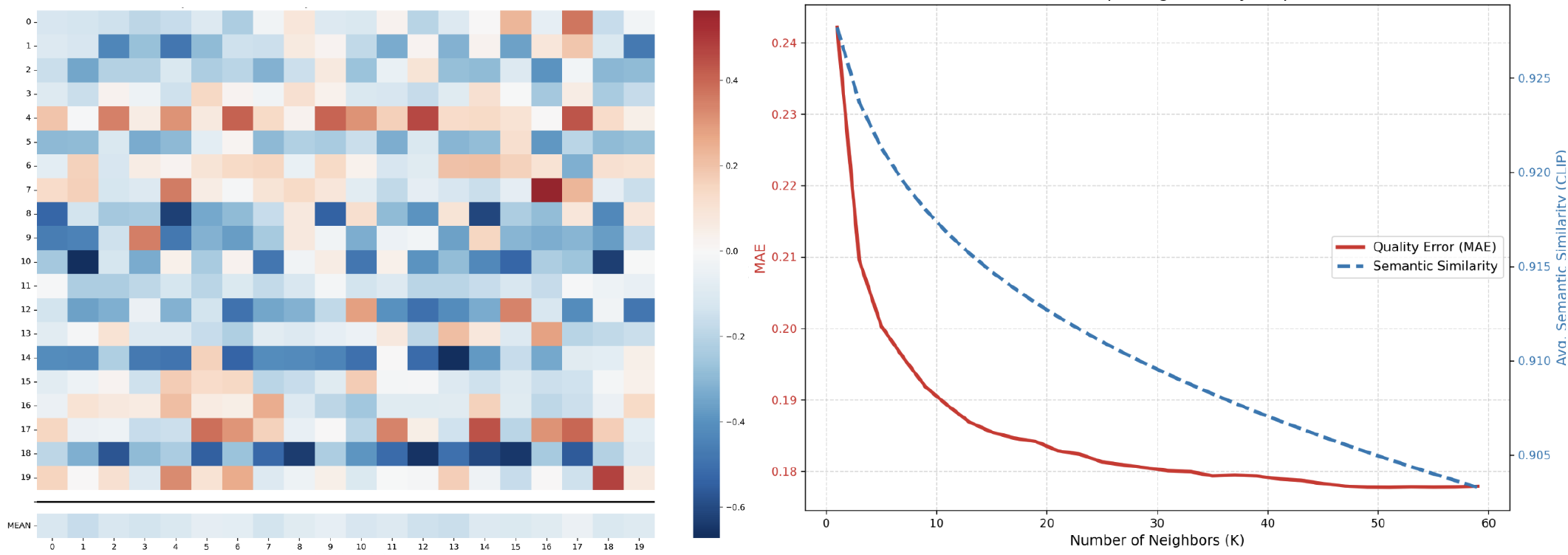}
    \caption{Visualization of retrieval consistency and the "Resonance" phenomenon. }
    % (Left) Deviation maps show that while individual nearest neighbors are noisy, their collective mean deviation converges toward zero as retrieval depth increases. (Right) The MAE continues to decrease even as semantic similarity declines, demonstrating that large-scale "fuzzy" consensus acts as a perceptual filter to cancel out content-driven noise.}
    \label{fig:res_exp}
\end{figure}
To justify the transition from single-neighbor retrieval to Fuzzy Quality Consensus, we visualize the consistency of retrieved neighbors in Fig.~\ref{fig:res_exp}. In Fig.~\ref{fig:res_exp} (left), each row corresponds to a query image, and each column denotes the deviation between the $k$-th neighbor and the ground truth score. The nearest neighbors often exhibit noticeable noise, indicating that the highest semantic similarity does not necessarily guarantee precise quality estimation. As the retrieval depth increases, however, the mean deviation steadily converges toward zero. This suggests that although individual memory anchors are noisy, their collective centroid provides a more stable calibration. Fig.~\ref{fig:res_exp} (right) reveals a counter-intuitive ``Resonance'' effect. As the number of neighbors $K$ increases, semantic similarity gradually declines, yet the quality error (MAE) continues to decrease and eventually stabilizes. This indicates that large-scale neighbor aggregation acts as a perceptual filter, suppressing content-driven noise and yielding a more consistent distortion representation. These observations support our central claim that fuzzy consensus over a broad cluster is more reliable than relying on a single highly similar but potentially biased neighbor. More qualitative experiments can be found in Sec.~\ref{sec:app_qua} of the Appendix.
% \begin{table}[h]
% \centering
% \caption{Performance comparison of synthetic reference galleries of different scales to authentic setting. Reference set as NONE indicates the intuitive score without using the reference gallery}
% \label{tab:gallery_ablation}
% \resizebox{0.7\columnwidth}{!}{
% \begin{tabular}{l cc cc cc}
% \toprule
% \multirow{2}{*}{Reference} 
% & \multicolumn{2}{c}{KONIQ} 
% & \multicolumn{2}{c}{LIVEC} & \multicolumn{2}{c}{BID}\\
% \cmidrule(lr){2-3} \cmidrule(lr){4-5} \cmidrule(lr){6-7}
% & SRCC & PLCC & SRCC & PLCC & SRCC & PLCC\\
% \midrule
% NONE &0.616  &0.653  &0.650  & 0.670 &0.689  & 0.711 \\
% LIVE &0.696  &0.707  & 0.707  & 0.742 & 0.731  & 0.747 \\
% CSIQ &0.693  &0.714 &0.723  &0.748 &0.735  &0.745 \\
% TID2013 &0.699  & 0.701  &0.728  & 0.751 & 0.735  & 0.745\\
% KADID &\textbf{0.765} & \textbf{0.759}&\textbf{0.778} &\textbf{0.780}  & \textbf{0.744} & \textbf{0.749}\\
% \bottomrule
% \end{tabular}
% }
% \end{table}

\subsection{Experiments on Robustness and Generalization}
We conducted a series of experiments to demonstrate the robustness and generalization ability of HIFC. More experiments can be found in Sec.~\ref{sec:app_rob} of the Appendix.

\textbf{Performance on Scale of Reference Datasets:} To analyze the impact of reference scale and diversity on cross-domain generalization, we replace the KADID gallery with smaller synthetic datasets. Results on three authentic benchmarks are summarized in Table~\ref{tab:gallery_ablation}.  Introducing even a small reference gallery (e.g., LIVE) consistently improves performance over the baseline without memory, demonstrating the effectiveness of fuzzy memory consensus. Moreover, performance increases progressively with the scale and diversity of the gallery. The largest gallery (KADID) achieves the best results, indicating that richer semantic contents and distortion types provide a denser and more representative anchor space for hierarchical retrieval, thereby enhancing robustness and accuracy.

% \begin{table}[t]
% \centering
% \caption{Cross-domain performance from synthetic to authentic datasets. Existing methods are trained with GT-MOS, while HIFC operates in a fully train-free and gt-free setting.}
% \label{tab:nogt}
% \resizebox{0.8\columnwidth}{!}{
% \begin{tabular}{c cc cc cc}
% \toprule
% \multirow{2}{*}{Ratio} 
% & \multicolumn{2}{c}{KONIQ} 
% & \multicolumn{2}{c}{LIVEC} & \multicolumn{2}{c}{BID}\\
% \cmidrule(lr){2-3} \cmidrule(lr){4-5} \cmidrule(lr){6-7}
% & SRCC & PLCC & SRCC & PLCC & SRCC & PLCC\\
% \midrule
% StyleAM & 0.584 & 0.561 & 0.700 & 0.673 & 0.637 & 0.567 \\
% FreqAlign & 0.618 & 0.588 & \textbf{0.748} & \underline{0.721} & 0.674 & 0.708  \\
% DGQA  & \underline{0.696} & \underline{0.690} & 0.681 & 0.687 & \textbf{0.770} & \textbf{0.753}\\
% Ours(w/o gt-mos) &\textbf{0.760}  & \textbf{0.751}  &\underline{0.745}&\textbf{0.734} & \underline{0.734}& \underline{0.733} \\
% \bottomrule
% \end{tabular}
% }
% \end{table}

\textbf{Experiments on Synthetic-to-Authentic without GT-MOS:} To assess the dependency on human annotations, we further evaluate a challenging variant in which the reference ground-truth MOS values are replaced by zero-shot intuition scores $S_{\text{int}}$ as pseudo-labels. As shown in Table~\ref{tab:nogt}, this annotation-free version still achieves highly competitive performance, reaching an SRCC of 0.760 on LIVEC, significantly outperforming UDA-based methods such as DGQA and StyleAM. Although a performance gap remains compared to the oracle version using ground-truth MOS, the overall robustness indicates that the proposed fuzzy memory consensus effectively mitigates individual label noise through cluster-based aggregation. These results confirm that HIFC can maintain accurate quality calibration even in the absence of precise human supervision.
\begin{table*}[tp]
\centering

\begin{minipage}[t]{0.48\textwidth}
\centering
\caption{Impact of reference gallery scale on cross-domain performance. ``NONE'' denotes inference without a memory bank.}
\label{tab:gallery_ablation}
\renewcommand{\arraystretch}{1.20}
\resizebox{\linewidth}{!}{
\begin{tabular}{l cc cc cc}
\toprule
\multirow{2}{*}{Reference} 
& \multicolumn{2}{c}{KONIQ} 
& \multicolumn{2}{c}{LIVEC} 
& \multicolumn{2}{c}{BID}\\
\cmidrule(lr){2-3} \cmidrule(lr){4-5} \cmidrule(lr){6-7}
& SRCC & PLCC & SRCC & PLCC & SRCC & PLCC\\
\midrule
NONE &0.616  &0.653  &0.650  &0.670 &0.689  &0.711 \\
LIVE &0.696  &0.707  &0.707  &0.742 &0.731  &0.747 \\
CSIQ &0.693  &0.714  &0.723  &0.748 &0.735  &0.745 \\
TID2013 &0.699  &0.701  &0.728  &0.751 &0.735  &0.745\\
KADID &\textbf{0.765} & \textbf{0.759}
      &\textbf{0.778} &\textbf{0.780}
      &\textbf{0.744} &\textbf{0.749}\\
\bottomrule
\end{tabular}
}
\end{minipage}
\hfill
\begin{minipage}[t]{0.48\textwidth}
\centering
\caption{Performance from synthetic to authentic datasets. Baselines are trained with GT-MOS supervision.}
\label{tab:nogt}
\renewcommand{\arraystretch}{1.15}
\resizebox{\linewidth}{!}{
\begin{tabular}{l cc cc cc}
\toprule
\multirow{2}{*}{Method} 
& \multicolumn{2}{c}{KONIQ} 
& \multicolumn{2}{c}{LIVEC} 
& \multicolumn{2}{c}{BID}\\
\cmidrule(lr){2-3} \cmidrule(lr){4-5} \cmidrule(lr){6-7}
& SRCC & PLCC & SRCC & PLCC & SRCC & PLCC\\
\midrule
StyleAM & 0.700 & 0.673  & 0.584 & 0.561  & 0.637 & 0.567 \\
FreqAlign & \textbf{0.748} & \underline{0.721}  & 0.618 & 0.588  & 0.674 & 0.708 \\
DGQA  & 0.681 & 0.687  & \underline{0.696} & \underline{0.690}  & \textbf{0.770} & \textbf{0.753}\\
Ours &\underline{0.745} &\textbf{0.734}
    &\textbf{0.760}  & \textbf{0.751}
     &\underline{0.734} &\underline{0.733} \\
\bottomrule
\end{tabular}
}
\renewcommand{\arraystretch}{1.0}
\end{minipage}
\end{table*}

\subsection{Ablation Studies}
\begin{table}[t]
\centering

\begin{minipage}[t]{0.49\linewidth}
\centering
\captionsetup{type=table}
\caption{Ablation study on different component combinations on the synthetic-to-authentic setting.}
\label{tab:ablation_components}

\resizebox{\linewidth}{!}{
\begin{tabular}{c ccc cccc}
\toprule
Index & SDR & IS & DA & LIVEC & KonIQ & BID & Avg \\
\midrule
a) & $\checkmark$ & & & 0.722 & 0.702 & 0.698 & 0.707 \\
b) & & $\checkmark$ & & 0.649 & 0.616 & 0.689 & 0.651 \\
c) & $\checkmark$ & $\checkmark$ & & 0.759 & 0.747 & 0.734 & 0.747 \\
d) & $\checkmark$ & & $\checkmark$ & 0.757 & 0.722 & 0.731 & 0.737 \\
\midrule
e) & $\checkmark$ & $\checkmark$ & $\checkmark$ & \textbf{0.778} & \textbf{0.765} & \textbf{0.744} & \textbf{0.762} \\
\bottomrule
\end{tabular}
}

\end{minipage}
\hfill
\begin{minipage}[t]{0.49\linewidth}
\centering
\captionsetup{type=table}
\caption{Ablation study on different feature space combinations of SDR. $\Rightarrow$ denotes coarse-to-fine retrieval.}
\label{tab:aba_squ}
\renewcommand{\arraystretch}{1.15}
\resizebox{\linewidth}{!}{
\begin{tabular}{c cc cc cc}
\toprule
\multirow{2}{*}{Space} 
& \multicolumn{2}{c}{LIVEC} 
& \multicolumn{2}{c}{KonIQ} 
& \multicolumn{2}{c}{BID} \\
\cmidrule(lr){2-3} \cmidrule(lr){4-5} \cmidrule(lr){6-7}
& SRCC & PLCC & SRCC & PLCC & SRCC & PLCC \\
\midrule
$f_{sem}$ & 0.733 & 0.731 & 0.667 & 0.688 & 0.730 & 0.736 \\
$f_{dis}$ & 0.713 & 0.721 & 0.714 & 0.704 & 0.627 & 0.632 \\
$f_{dis} \Rightarrow f_{sem}$ & 0.770 & 0.772 & 0.757 & 0.750 & 0.727 & 0.729 \\
$f_{sem} \Rightarrow f_{dis}$ & \textbf{0.778} & \textbf{0.780} & \textbf{0.765} & \textbf{0.759} & \textbf{0.744} & \textbf{0.749} \\
\bottomrule
\end{tabular}
}
\renewcommand{\arraystretch}{1.0}
\end{minipage}
\end{table}
\textbf{Ablation Studies on Components:} We analyze the contributions of Intuition Score (IS), Distortion Alignment (DA), and Semantic-Distortion Retrieval (SDR) in Table~\ref{tab:ablation_components}. As shown in rows (a), (b), and (e), neither SDR nor IS alone achieves optimal performance, while their combination (row e) yields the best average SRCC of 0.762, highlighting the importance of integrating intuition and memory. The performance gain also verifies the role of DA, which aligns distortion manifolds between synthetic and authentic domains, enabling more reliable neighbor retrieval. Furthermore, row (d) shows that removing IS from the full memory pipeline (SDR+DA) degrades performance, indicating that IS provides a crucial zero-shot baseline that stabilizes overall quality prediction.
% We evaluate the individual contributions of the Intuition Score (IS), Distortion Alignment (DA), and Semantic-Distortion Retrieval (SDR) in Table~\ref{tab:ablation_components}. Comparing rows (a), (b), and (e), neither the memory-based SDR nor the intuition-based IS alone achieves optimal results.  Their integration (row e) yields the best average SRCC of 0.762, confirming that the synergy between "affective intuition" and "rational memory" is essential for robust IQA. This performance gain validates the necessity of cross-domain distortion type alignment. By explicitly aligning the distortion manifolds between the synthetic reference and authentic target domains, the model can more accurately locate perceptually consistent neighbors in the memory bank.  Row (d) shows that even with a complete memory path (SDR+DA), omitting the intuition score (IS) leads to a performance drop.  This underscores that IS provides a necessary zero-shot baseline that stabilizes the overall quality calibration.

\textbf{Ablation Study on Component of Semantic-Distortion Retrieval:} We analyze the design of the Semantic-Distortion Retrieval (SDR) module by comparing different retrieval strategies. As shown in Table~\ref{tab:aba_squ}, using a single feature space, either semantic ($f_{\text{sem}}$) or distortion-aware ($f_{\text{dis}}$)—leads to suboptimal results, as neither alone ensures both content relevance and distortion consistency.
In contrast, the proposed semantic-first, distortion-second strategy ($f_{\text{sem}} \Rightarrow f_{\text{dis}}$) consistently achieves the best performance, outperforming the reverse order. This indicates that semantic filtering provides a stable context, enabling more precise distortion-based re-ranking. Such hierarchical design improves retrieval reliability and leads to more accurate quality estimation. Additional ablation results and analyses are provided in the Appendix ~\ref{sec:app_aba}.
% We further analyze the design of the Semantic-Distortion Retrieval (SDR) module by comparing different retrieval and re-ranking strategies. As reported in Table~\ref{tab:aba_squ}, relying on a single feature space, either purely semantic ($f_{\text{sem}}$) or purely distortion-aware ($f_{\text{dis}}$), leads to suboptimal performance, as neither strategy alone can simultaneously ensure content relevance and degradation consistency. In contrast, the proposed \emph{Semantic-first, Distortion-second} strategy ($f_{\text{sem}} \Rightarrow f_{\text{dis}}$) consistently achieves the best results across all evaluated datasets, significantly outperforming the reverse order ($f_{\text{dis}} \Rightarrow f_{\text{sem}}$). This observation suggests that constraining the search space to semantically similar scenes provides a stable contextual prior, upon which fine-grained distortion re-ranking can more precisely capture quality-related differences. Such hierarchical synergy enables the model to identify the most relevant memory anchors, thereby maximizing the reliability and accuracy of the final quality consensus.

\textbf{Contribution beyond the CLIP backbone}: To further evaluate the effectiveness of HIFC-IQA, we retrain DGQA and SynDR-IQA using the same ViT-B/32 backbone and KADID-10k source dataset. As shown in Table~\ref{tab:same-backbone-methods}, HIFC-IQA achieves the best performance on LIVEC and KonIQ-10k, reaching SRCC/PLCC values of 0.778/0.780 and 0.765/0.759, respectively, while maintaining comparable performance on BID.
Notably, HIFC-IQA achieves these results without IQA-specific training, demonstrating the effectiveness of its hierarchical retrieval and fuzzy quality consensus mechanisms in leveraging frozen visual representations.

\begin{table}[h]
\centering
\caption{Comparison with DGQA and SynDR-IQA retrained using the same ViT-B/32 backbone. The source dataset/reference gallery is KADID-10k. }
\label{tab:same-backbone-methods}
\small
\begin{tabular}{@{}lcccc@{}}
\toprule
Method & IQA train & LIVEC & KonIQ-10k & BID \\
\midrule
DGQA (ViT-B/32) & Yes & 0.712/0.721 & 0.697/0.701 & 0.723/0.731 \\
SynDR-IQA (ViT-B/32) & Yes & 0.732/0.736 & 0.752/0.748 & 0.745/0.754 \\
HIFC-IQA & No & 0.778/0.780 & 0.765/0.759 & 0.744/0.749 \\
\bottomrule
\end{tabular}
\end{table}

\textbf{Gallery size and distortion diversity}: We investigate the influence of reference gallery size by varying the proportion of KADID-10k samples from 10\% to 100\%.
As shown in Table~\ref{tab:gallery-subsets}, increasing the gallery size consistently improves prediction performance across both target datasets.
Specifically, the SRCC increases from 0.734 to 0.778 on LIVEC and from 0.729 to 0.765 on KonIQ-10k. These results indicate that a larger reference gallery provides richer quality exemplars, facilitating more reliable neighbor retrieval and fuzzy quality consensus estimation.

\begin{table}[h]
\centering
\caption{Effect of KADID-10k gallery size on transfer to LIVEC and KonIQ-10k.}
\label{tab:gallery-subsets}
\small
\begin{tabular}{@{}lcc@{}}
\toprule
KADID-10k subset & LIVEC & KonIQ-10k \\
\midrule
10\% & 0.734/0.746 & 0.729/0.727 \\
25\% & 0.741/0.752 & 0.752/0.750 \\
50\% & 0.756/0.766 & 0.755/0.755 \\
100\% & 0.778/0.780 & 0.765/0.759 \\
\bottomrule
\end{tabular}
\end{table}

\textbf{Effect of Distortion Diversity.} We further investigate the impact of distortion diversity by fixing the gallery size at 2,000 images while varying the number of distortion types.
As shown in Table~\ref{tab:distortion-diversity}, the best performance on both datasets is achieved with 10 distortion types, while further increasing diversity leads to performance degradation. This observation suggests a trade-off between distortion coverage and the density of quality exemplars within each distortion category.
Under a fixed gallery capacity, a balanced distortion distribution enables more effective local quality matching and consensus estimation.

\begin{table}[h]
\centering
\caption{Effect of distortion diversity at a fixed gallery size of 2,000 images.}
\label{tab:distortion-diversity}
\small
\begin{tabular}{@{}lcc@{}}
\toprule
Distortion diversity & LIVEC & KonIQ-10k \\
\midrule
5 types & 0.745/0.739 & 0.604/0.644 \\
10 types & 0.767/0.764 & 0.612/0.647 \\
20 types & 0.735/0.746 & 0.602/0.642 \\
Full diversity & 0.735/0.745 & 0.602/0.642 \\
\bottomrule
\end{tabular}
\end{table}

\section{Conclusion}
In this paper, we propose HIFC, a train-free framework that fundamentally rethinks Cross-Domain Image Quality Assessment (CD-IQA). Inspired by the dual-process mechanism of the human visual system, HIFC synergizes zero-shot intuitive perception with a structured fuzzy memory consensus. By decoupling physical distortions from semantic interference, our framework establishes robust perceptual anchors across diverse manifolds. Extensive experiments confirm that without relying on target-domain pseudo-labeling or costly parameter updates, HIFC achieves highly competitive generalization, particularly in the challenging synthetic-to-authentic setting, rivaling state-of-the-art domain adaptation methods. Ultimately, HIFC highlights the immense potential of biologically-inspired, plug-and-play paradigms for robust real-world visual quality evaluation.

% \section*{References}

% \medskip

{
\small

\bibliographystyle{plainnat}  % 或其他 natbib 兼容样式，如 abbrvnat, unsrtnat
\bibliography{references}     % 对应的 .bib 文件名（不含扩展名）
}

%%%%%%%%%%%%%%%%%%%%%%%%%%%%%%%%%%%%%%%%%%%%%%%%%%%%%%%%%%%%
\newpage
\appendix

\section{Introduction to the Appendix}
\label{sec:app_intro}
This appendix provides additional technical details, theoretical analysis, and extended experimental results to further substantiate the findings of the main paper. We aim to enhance reproducibility, deepen theoretical understanding, and present broader empirical evidence of the effectiveness and robustness of the proposed HIFC framework. Specifically, this Appendix includes:

\begin{itemize}
    \item \textbf{Additional Experimental Setups.} Detailed implementation specifications of HIFC, including backbone configurations, hyperparameter settings, and dataset-specific evaluation protocols.
    
    \item \textbf{Theoretical Analysis of Memory Quality Consensus.} A rigorous mathematical formulation of the Memory Quality Consensus mechanism, clarifying how fuzzy aggregation suppresses individual anchor bias and stabilizes memory-based calibration.
    
    \item \textbf{Extended Ablation Studies.} Further investigations into critical design choices, such as patch granularity in the Stochastic Jigsaw module and additional hyper-parameter analysis of the framework.
    
    \item \textbf{Additional Qualitative Analysis.} Supplementary visualizations, including extended gMAD comparison pairs and retrieval visualization examples, to better illustrate perceptual alignment.
    
    \item \textbf{Robustness and Generalization Experiments.} Additional stress tests under extreme degradation conditions.

    \item \textbf{Additional Cross-Domain and Few-Shot Generalization Experiments.} Further cross-domain evaluations on non-traditional distortions and aesthetic benchmarks, along with a few-shot style comparison by varying the reference gallery source.
    
    \item \textbf{Failure Case Analysis.} Representative examples where HIFC underperforms, along with diagnostic discussions on the limitations of the current hierarchical retrieval mechanism.
    
    \item \textbf{Limitations and Future Directions.} A discussion of the boundaries of the training-free paradigm and potential future extensions, including enriching the rational memory path with more diverse modalities.
\end{itemize}

\section{Additional Experimental Setups}
\label{sec:add_exp_setup}
\subsection{Additional Implementation Details}
\noindent\textbf{Computational Environment.}
To ensure reproducible and efficient inference, the structured reference gallery is indexed using the FAISS library~\cite{faiss}. All experiments are conducted on an NVIDIA RTX 4090 GPU. The proposed HIFC framework achieves an average inference latency of 14 ms per image, meeting the requirements for real-time industrial deployment. To eliminate the random seed bias, we conducted ten experiments with different seeds and took the final median as the experimental result.

\noindent\textbf{Preprocessing Rationales.}
The stride of the semantic branch is set to $224$ to match the encoder input resolution and avoid redundant feature extraction. For the Scrambled Jigsaw, the grid size is chosen as $G_{\text{size}} = 32 / 2 = 16$, which maximizes semantic disruption while preserving fine-grained pixel-level degradation characteristics.

\noindent\textbf{Hyperparameters.}
The temperature scaling factor in the polarization path is set to $\tau = 100$. The reference gallery $\mathcal{G}$ is constructed from the KADID-10k~\cite{kadid-10k} dataset. All features are $\ell_2$-normalized prior to indexing to ensure consistency with cosine similarity during retrieval.

\subsection{Evaluation Metrics}
In image quality assessment (IQA), correlation-based metrics are widely used to evaluate the relationship between predicted quality scores and human-perceived scores. Two commonly used metrics in this context are the Spearman Rank Correlation Coefficient (SRCC) and the Pearson Linear Correlation Coefficient (PLCC). These metrics are employed to assess the monotonicity and linearity of the relationship between predicted and ground truth quality scores, respectively.

\textbf{Spearman's Rank Correlation Coefficient (SRCC)} : Spearman's Rank Correlation Coefficient (SRCC) measures the monotonic relationship between two variables, assessing how well the relationship between two variables can be described using a monotonic function. Unlike Pearson's correlation, SRCC does not require the variables to be linearly related and is thus more robust to outliers.
Given a set of ground truth quality scores \( y = \{ y_1, y_2, \dots, y_n \} \) and predicted quality scores \( \hat{y} = \{ \hat{y}_1, \hat{y}_2, \dots, \hat{y}_n \} \), the SRCC is computed as follows:

\begin{equation}
SRCC
= 1 - \frac{6 \sum_{i=1}^{n} d_i^2}{n(n^2 - 1)}
\end{equation}
where:
\begin{itemize}
    \item \( d_i = \hat{r}_i - r_i \) is the difference between the rank of the predicted score \( \hat{y}_i \) and the rank of the ground truth score \( y_i \),
    \item \( r_i \) is the rank of the ground truth score \( y_i \),
    \item \( \hat{r}_i \) is the rank of the predicted score \( \hat{y}_i \),
    \item \( n \) is the number of data points.
\end{itemize}

\textbf{Pearson Linear Correlation Coefficient (PLCC) : }The Pearson Linear Correlation Coefficient (PLCC) is used to assess the linear relationship between two variables. It quantifies how well the predicted scores and the ground truth scores are linearly related, assuming that the relationship between the two variables is linear.

For the ground truth scores \( y = \{ y_1, y_2, \dots, y_n \} \) and predicted scores \( \hat{y} = \{ \hat{y}_1, \hat{y}_2, \dots, \hat{y}_n \} \), PLCC is computed as follows:

\begin{equation}
\text{PLCC}(y, \hat{y}) = \frac{\sum_{i=1}^{n} (y_i - \bar{y})(\hat{y}_i - \bar{\hat{y}})}{\sqrt{\sum_{i=1}^{n} (y_i - \bar{y})^2 \sum_{i=1}^{n} (\hat{y}_i - \bar{\hat{y}})^2}}
\end{equation}

where:
\begin{itemize}
    \item \( \bar{y} \) and \( \bar{\hat{y}} \) are the mean values of the ground truth scores and predicted scores, respectively:
    \[
    \bar{y} = \frac{1}{n} \sum_{i=1}^{n} y_i, \quad \bar{\hat{y}} = \frac{1}{n} \sum_{i=1}^{n} \hat{y}_i,
    \]
    \item \( n \) is the number of data points.
\end{itemize}

\section{Theoretical Analysis of Memory Quality Consensus}
\label{sec:MQC}

Conventional Retrieval-Augmented Generation (RAG) paradigms in Image Quality Assessment (IQA) typically follow a point-wise precise matching strategy. However, human memory of image quality is inherently fuzzy and prototype-based. To bridge the gap between feature space representations and human subjective perception, we propose a HIFC framework. By integrating the trace-retention property of visual encoders with the Aristotelian Representation Hypothesis~\cite{Aristotelian}, we theoretically justify the optimality and robustness of the \emph{large-neighborhood retrieval centroid} as a fuzzy memory representation.

\subsection{Underlying Support of the Quality Manifold: Processing Traces in Visual Encoders}

Traditional perspectives assume that pre-trained visual models (e.g., CLIP) primarily encode high-level semantic information and are insensitive to low-level distortions such as blur, noise, or compression artifacts. However, recent findings by Ramos et al.~\cite{ramos2025processing} reveal that the latent space of visual encoders systematically preserves \emph{Processing Traces} and \emph{Acquisition Traces}.

Within the feature space $\mathcal{Z}$, any image representation $z_i$ can be viewed as an entangled composition of semantic content features $z_{\text{content}}$ and quality-related degradation traces $z_{\text{trace}}$. Because common distortions (e.g., Gaussian blur or JPEG compression) exhibit strong statistical regularities at low levels, these traces are not randomly scattered in latent space. Instead, they spontaneously cluster into distinct local quality sub-manifolds. 

This phenomenon provides both physical and representational justification for capturing image quality through clustering or neighborhood retrieval. Quality attributes do not correspond to isolated points, but rather to structured regions with well-defined topology in latent space.

\subsection{Local Consistency Guarantee: The Aristotelian Representation Hypothesis}

Although degradation traces form sub-manifolds, global coordinate systems across different models (or even different scales of the same model) exhibit significant scale bias. Gröger et al.~\cite{Aristotelian} propose the \emph{Aristotelian Representation Hypothesis}, which states that neural network representations do not converge toward a single global Platonic ideal space~\cite{huh2024position}, but instead converge toward shared local neighborhood relationships.

Let $\mathcal{N}_k(z_q)$ denote the $k$-nearest neighbors of a query feature $z_q$ in latent space. According to the Aristotelian hypothesis, although global distance distributions may differ across models $\mathcal{M}_A$ and $\mathcal{M}_B$, their local geometric relationships remain approximately isomorphic:

\[
\text{Rank}_{\mathcal{M}_A}(\mathcal{N}_k(z_q)) 
\approx 
\text{Rank}_{\mathcal{M}_B}(\mathcal{N}_k(z_q)).
\]

This implies that large-neighborhood retrieval yields a quality-consistent sample set with strong cross-model and cross-scale robustness. Rather than relying on unstable global coordinates, we anchor quality estimation on locally agreed manifold structures.

\subsection{Centroid Consensus Convergence under Fuzzy Memory Mechanism}

Prototype Theory in cognitive psychology~\cite{rosch2024principles} suggests that humans rely on category prototypes, the central statistical tendencies when making judgments under uncertain or fuzzy memory conditions.

Within the hierarchical memory framework, the retrieved $k$-nearest samples are interpreted as the set of recalled experiences activated in fuzzy memory. Individual samples inevitably contain semantic noise and idiosyncratic bias. Directly relying on Top-1 retrieval or discrete Top-$k$ aggregation (as in traditional RAG~\cite{lewis2020retrieval,asai2023self}) leads to high-variance estimations.

To address this issue, we compute the centroid over the large neighborhood $\mathcal{N}_k(z_q)$ to extract a quality consensus vector $c_q$:

\[
c_q = \mathbb{E}_{z \sim \mathcal{N}_k(z_q)}[z]
\approx
\frac{1}{k} \sum_{i=1}^{k} z_i.
\]

From the perspective of statistical learning theory, the centroid $c_q$ is the optimal estimator minimizing the total mean squared error within the neighborhood:

\[
c_q = \arg\min_{c} \sum_{z_i \in \mathcal{N}_k(z_q)} \| z_i - c \|^2.
\]

\paragraph{Theoretical Implications.}

\textbf{1) Denoising Effect.}  
The large-neighborhood centroid operation acts as a low-pass filter over the quality sub-manifold. It suppresses semantic variance among neighboring samples and purifies the homogeneous processing traces identified by Ramos et al.~\cite{ramos2025processing}.

\textbf{2) Representation Alignment.}  
Combined with the Aristotelian hypothesis~\cite{Aristotelian}, the stability of local neighborhoods ensures that the centroid corresponds to a robust Aristotelian quality prototype at a given degradation level. This mechanism closely simulates human perception, where detailed content is abstracted away and overall visual quality is judged through stable consensus.

In summary, unlike traditional RAG frameworks that pursue information completion, our large-neighborhood centroid retrieval extracts a purified \emph{quality-trace prototype} from structured local manifolds. Theoretically, this mechanism exhibits stronger cognitive plausibility and superior cross-domain robustness.

\section{More Ablation Experiments}
\label{sec:app_aba}
\subsection{Ablation Study on Scale of Pre-trained Models}
We examine the sensitivity of HIFC to different pre-trained CLIP backbones, including convolutional architectures (RN50) and Vision Transformers of varying scales (ViT-B/32, ViT-B/16, and ViT-L/14). The comparative results are reported in Table~\ref{tab:backbone_ablation}. HIFC maintains strong and stable performance across all tested backbones, demonstrating its robustness without requiring architecture-specific adjustments. Notably, the performance does not increase monotonically with model capacity. Although ViT-B/32 achieves the best overall results, the substantially larger ViT-L/14 shows a clear performance drop. We attribute this trend to the trade-off between semantic abstraction and distortion sensitivity. Larger models offer stronger semantic representation ability but tend to overemphasize high-level content semantics, potentially suppressing subtle low-level artifacts that are critical for image quality assessment. These findings suggest that, within our hierarchical retrieval framework, a medium-scale backbone such as ViT-B/32 strikes the most effective balance, preserving distortion-relevant details while maintaining a reliable semantic prior for memory consensus.
\begin{table}[h]
\centering
\caption{Performance comparison of backbones at different scales.}
\label{tab:backbone_ablation}
\resizebox{0.6\columnwidth}{!}{
\begin{tabular}{c cc cc cc}
\toprule
\small
\multirow{2}{*}{Models} 
& \multicolumn{2}{c}{LIVEC} 
& \multicolumn{2}{c}{KONIQ}\\
\cmidrule(lr){2-3} \cmidrule(lr){4-5}
& SRCC & PLCC & SRCC & PLCC \\
\midrule
RN50 & 0.700 & 0.705  &0.693  &0.707   \\
B-32 & \textbf{0.778} &\textbf{0.780} &0.765  &\textbf{0.759 } \\
B-16 & 0.718 & 0.750 &\textbf{0.770} &0.756\\
L-14 & 0.710 & 0.722  &0.731  &0.750   \\
\bottomrule
\end{tabular}
}
\end{table}

\begin{figure}[t]
    \centering
    \includegraphics[width=\linewidth]{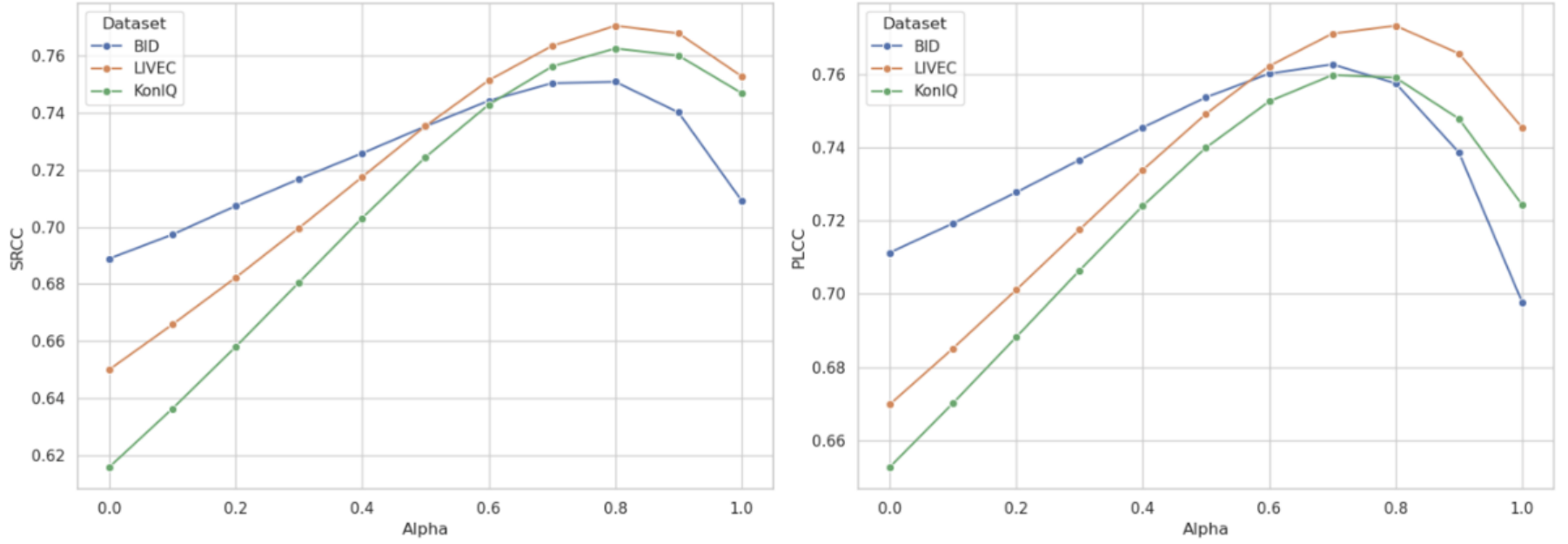}
    \caption{Sensitivity analysis of the fusion factor $\alpha$ on three authentic datasets. The results show a consistent concave trend with an optimal peak around $\alpha=0.8$, demonstrating that the rational memory consensus serves as the primary calibration source while affective intuition provides essential perceptual regularization. The performance degradation at both extremes validates that the synergy between intuition and memory is superior to relying on either component in isolation.}
    \label{fig:aba_alpha}
\end{figure}

\subsection{Ablation Studies on Integration Coefficient $\alpha$}
We perform a sensitivity analysis on the fusion factor $\alpha$, which controls the balance between the affective intuition score ($S_{\text{int}}$) and the rational memory consensus ($S_{\text{mem}}$). The SRCC and PLCC results on three authentic datasets are presented in Fig.~\ref{fig:aba_alpha}.

As shown in the figure, all datasets exhibit a consistent concave performance trend, with the optimal point located around $\alpha = 0.8$. This observation provides two important insights. The peak occurring at a relatively high $\alpha$ indicates that the hierarchical retrieval and fuzzy consensus mechanism ($S_{\text{mem}}$) serves as the primary and most reliable source of quality calibration for unseen authentic distortions. The noticeable performance drop at $\alpha = 1.0$, where only $S_{\text{mem}}$ is used, reveals that zero-shot affective intuition plays a crucial regularization role. It acts as a perceptual anchor that mitigates potential bias introduced by outlier memory neighbors.

The significant degradation at both extremes ($\alpha = 0$ and $\alpha = 1$) further validates the central philosophy of HIFC: robust blind quality assessment emerges from the synergy between rapid intuition and deliberative memory, rather than relying exclusively on either component.

\subsection{Additional Hyper-parameter Analysis}
Analogous to Retrieval-Augmented Generation (RAG) frameworks~\cite{lewis2020retrieval,asai2023self,edge2404local}, HIFC demonstrates strong adaptability to diverse tasks through lightweight hyperparameter calibration. While the main paper reports results under a unified parameter setting, we further explore task-specific optimization of $N_{\text{dist}}$ (number of distortion prototypes) and $k$ (Top-$k$ retrieval depth). The refined results are summarized in Table~\ref{tab:hypers}. After tuning hyperparameters for individual datasets, HIFC achieves further performance improvements. For example, the SRCC on LIVEC and BID increases to 0.785 and 0.764, respectively. This plug-and-play flexibility suggests that HIFC can function as a foundation model for IQA, where the internal memory consensus can be adaptively aligned with the distortion distribution of a target domain.

We also analyze the relationship between these hyperparameters and the statistical properties of the distortion-aware cosine similarity manifold. As shown in the Mean and Std columns of Table~\ref{tab:hypers}, two consistent trends emerge:

\emph{Diversity Scaling} ($N_{\text{dist}} \propto \text{Std}$):  
The standard deviation of the distortion-centroid similarity is positively correlated with $N_{\text{dist}}$. Increasing the number of distortion prototypes enlarges the diversity of the alignment space, enabling more fine-grained discrimination of complex artifacts.

\emph{Confidence Scaling} ($k \propto \text{Mean}$):  
The retrieval depth $k$ shows a positive correlation with the mean similarity score. A larger retrieval pool forms a denser consensus manifold within the memory bank, improving the stability and confidence of the final quality estimation.

These observations indicate that the effectiveness of HIFC is not merely the result of empirical parameter selection, but is rooted in its ability to preserve a statistically stable and interpretable alignment manifold across unseen domains.

\begin{table}[h]
\centering
\caption{Hyperparameter analysis and statistical correlation. The results reveal that $N_{dist}$ scales with similarity variance to enhance distortion diversity, while k correlates with mean similarity to improve consensus confidence. }
\label{tab:hypers}
\resizebox{0.7\columnwidth}{!}{
\begin{tabular}{c cc cc cc}
\toprule
Dataset & $N_{dist}$ & TOPK & Mean & Std & PLCC & SRCC\\
\midrule
KONIQ&4&400&0.893&0.006&0.783 & 0.774\\
LIVEC&12&400&0.892&0.008&0.784 & 0.785 \\
BID&16&700&0.903&0.009&0.758 &0.764\\
PIPAL&4&500&0.900&0.006&0.508 &0.513\\
\bottomrule
\end{tabular}
}
\end{table}

\section{Additional Qualitative Analysis}
\label{sec:app_qua}

\subsection{Additional gMAD experiment with CLIP-IQA}
\begin{figure}[h]
    \centering
    \includegraphics[width=\linewidth]{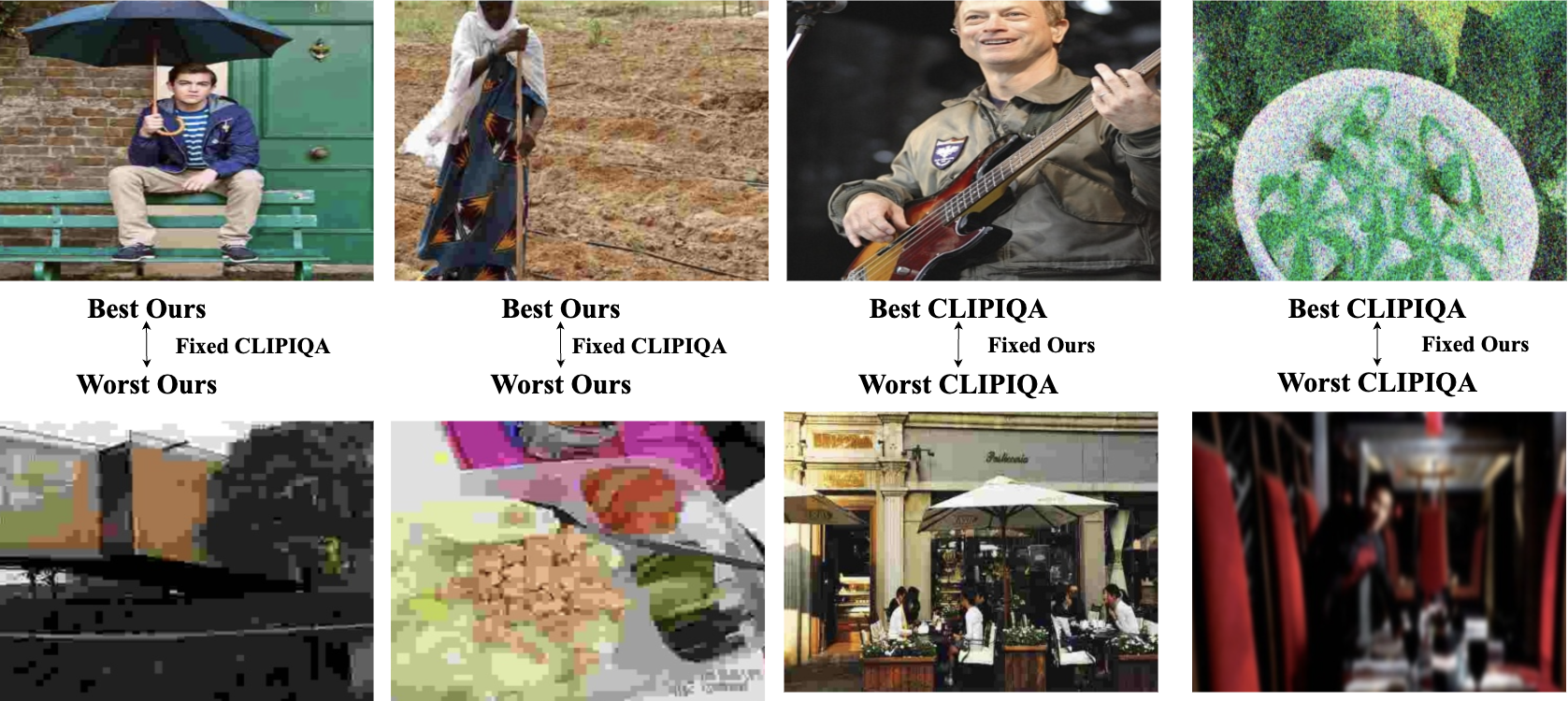}
    \caption{Representative gMAD pairs between HIFC and CLIP-IQA on synthetic distortions.}
    \label{fig:Gmad_CLIPiqa}
\end{figure}

To intuitively evaluate the perceptual alignment and robustness of our proposed framework, we conduct a group Maximum Differentiation (gMAD) competition~\cite{gmad} between HIFC and CLIP-IQA~\cite{clipiqa}. Fig.~\ref{fig:Gmad_CLIPiqa} illustrates the gMAD pairs generated by the competition. In the right column, where HIFC acts as the defender and CLIP-IQA as the attacker, the paired images exhibit highly similar visual quality levels despite containing different content and degradation types. This indicates that HIFC successfully resists the attack from CLIP-IQA, maintaining stable and accurate predictions. Conversely, when the roles are reversed and CLIP-IQA acts as the defender, HIFC consistently identifies the severe failure cases of CLIP-IQA. The gMAD experiment provides an intuitive proof that the capabilities of HIFC do not merely stem from the powerful visual language Encoder. This further validates the effectiveness of our approach.\\

\subsection{Additional gMAD experiment with DGQA}
\begin{figure}[h]
    \centering
    \includegraphics[width=\linewidth]{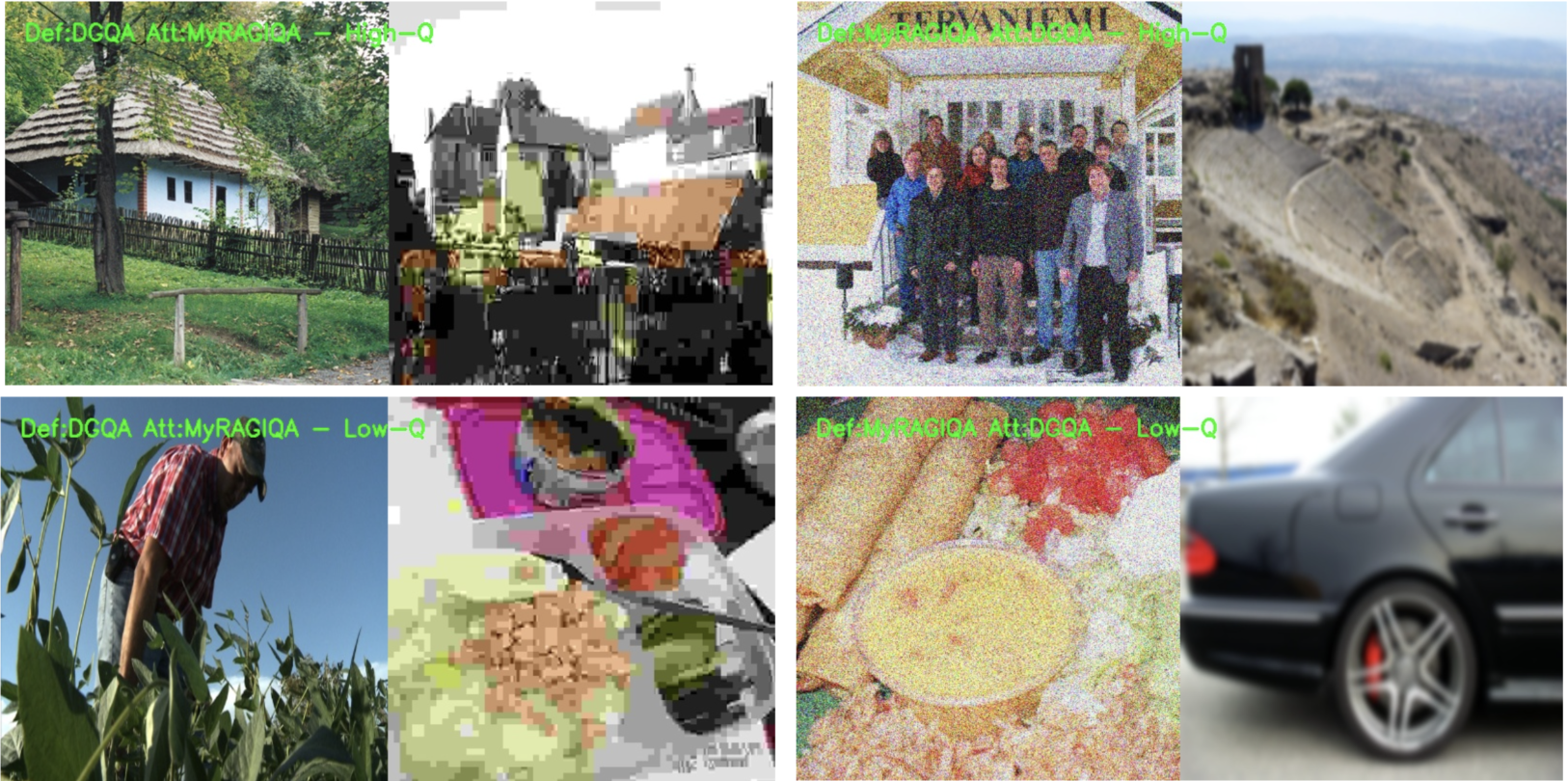}
    \caption{Representative gMAD pairs between HIFC and DGQA.}
    \label{fig:gmad_dgqa}
\end{figure}

To intuitively evaluate the perceptual alignment and robustness of our proposed framework, we conduct a group Maximum Differentiation (gMAD) competition~\cite{gmad} between HIFC and DGQA. Fig.~\ref{fig:gmad_dgqa} illustrates the gMAD pairs generated by the competition. In the right column, where HIFC acts as the defender and DGQA as the attacker, the paired images exhibit highly similar visual quality levels despite containing different content and degradation types. This indicates that HIFC successfully resists the attack from DGQA, maintaining stable and accurate predictions. Conversely, when the roles are reversed and DGQA acts as the defender (left column), HIFC consistently identifies the severe failure cases of DGQA which display substantial differences in quality according to human perception, pairing relatively clear, high-quality scenes with heavily distorted, blocky artifacts, yet DGQA incorrectly rates them as having the same quality. These qualitative findings strongly align with our quantitative results, suggesting that by leveraging the synergy of affective intuition and memory consensus, HIFC generalizes significantly better to complex, unseen distortions and aligns more closely with the human visual system than heavily optimized training-based models.

\subsection{Additional TSNE Analysis}
\begin{figure}[h]
    \centering
    \includegraphics[width=\linewidth]{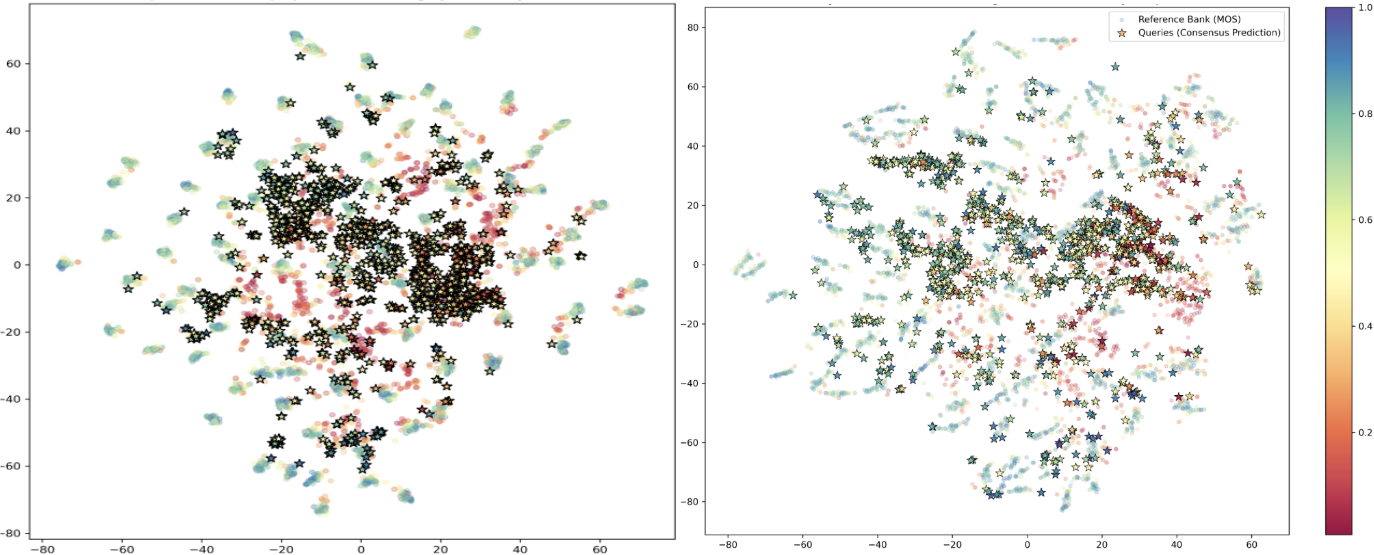}
    \caption{t-SNE visualization of the feature manifold for the reference bank and query images. The left panel shows that the distortion-aware feature space effectively suppresses semantic interference to form distinct degradation clusters, where query images (stars) are accurately embedded among relevant prototypes. The right panel displays a globally continuous quality gradient, confirming that the Quality-Centroid Consensus achieves robust perceptual alignment by performing localized calibration within a structured and consistent manifold.}
    \label{fig:tsne}
\end{figure}

To provide a deeper insight into the structure of the learned representation and the reliability of our retrieval mechanism, we visualize the feature manifold of the reference bank and query images using t-SNE~\cite{tsne} in Fig.~\ref{fig:tsne}.

Distortion-Discriminative Embedding (Left). In the left panel, the reference bank is colored by distortion types, while query images (represented by stars) are projected into the same space. The clear, localized clustering of different colors demonstrates that our feature space refined by Distortion Alignment (DA) effectively suppresses semantic interference and organizes images based on their underlying degradation characteristics. The query stars are correctly embedded within the clusters of relevant reference degradations, providing a solid foundation for subsequent hierarchical retrieval.

Quality-Consistent Manifold (Right). The right panel colors the reference bank by ground-truth MOS and the query images by their Consensus Predictions. We observe a globally continuous color gradient across the manifold, transitioning smoothly from low-quality (warm colors) to high-quality (cool colors) regions. Crucially, the predicted scores of the query stars are highly consistent with the colors of their immediate reference neighbors. This visual evidence validates the efficacy of the Quality-Centroid Consensus (QCC): the "fuzzy" consensus is not a simple average of random samples, but a localized calibration within a perceptually structured manifold. By retrieving and aggregating from this consistent space, HIFC effectively neutralizes idiosyncratic noise and achieves robust alignment with human perception.
\subsection{Additional Visualization Analysis}
\begin{figure}[h]
    \centering
    \includegraphics[width=\linewidth]{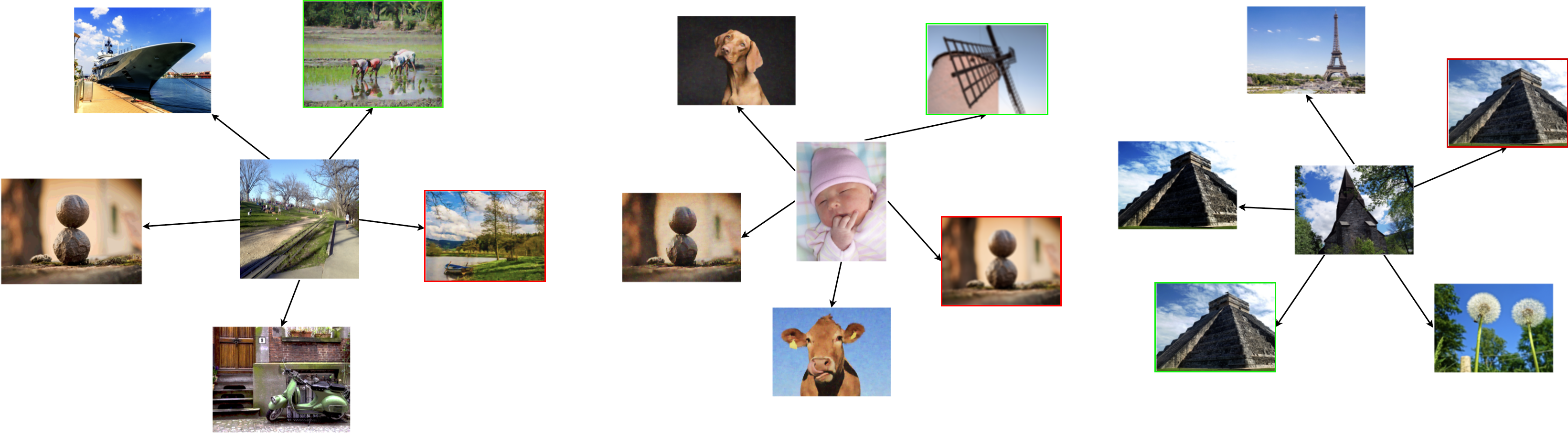}
    \caption{Qualitative visualization of the top-5 neighbors supporting the Fuzzy Quality Consensus.}
    \label{fig:vis_ret}
\end{figure}

To further illustrate the stability of the Fuzzy Quality Consensus, we visualize the top-5 neighbors that most strongly support the retrieved quality centroid for three representative queries in Fig.~\ref{fig:vis_ret}. These neighbors are selected based on their proximity to the final aggregated quality score within the memory manifold. Across all three examples, the retrieved neighbors demonstrate strong perceptual consistency with their corresponding queries. In the left-most group, although the semantic categories range from ships to urban parks, the neighbors share highly similar authentic degradation patterns, such as subtle motion blur and lighting artifacts. This observation confirms that the proposed Semantic-Distortion Retrieval (SDR) mechanism effectively emphasizes distortion characteristics, while leveraging semantic similarity as a stable initialization prior. The visualization further shows that the quality centroid is not a simple statistical average of arbitrary samples, but a perceptual anchor supported by a coherent cluster of neighbors. Even when noticeable semantic gaps exist—for example, the baby query in the middle group being matched with animal or architectural scenes—the overall quality manifold remains well aligned. 

These qualitative results reinforce the central philosophy of HIFC: by forming consensus over a retrieved cluster rather than relying on a single nearest neighbor, the framework mitigates content-specific bias and more reliably captures the underlying distortion essence of the image.

\subsection{Visualization Analysis of Scrambled Jigsaw Puzzle}
\begin{figure}[h]
    \centering
    \includegraphics[width=\linewidth]{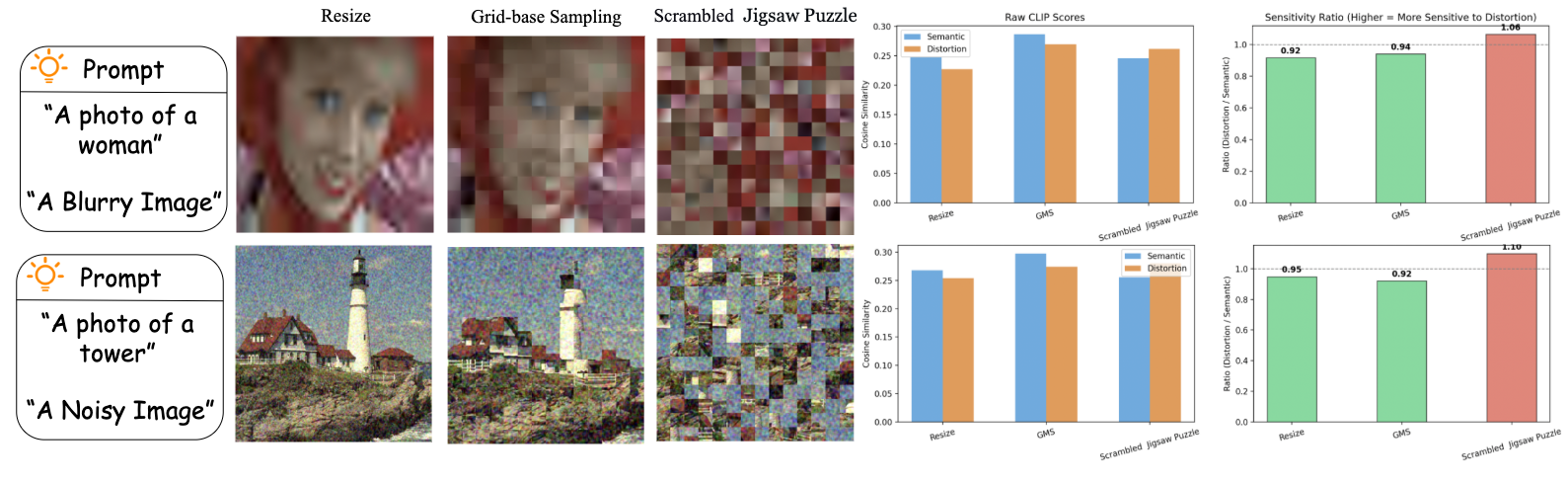}
    \caption{Qualitative and quantitative comparison of different preprocessing strategies.}
    \label{fig:aba_gms}
\end{figure}

We present both qualitative and quantitative comparisons of different preprocessing strategies—Resize, Grid-based Sampling (GMS)~\cite{fastvqa}, and the proposed Scrambled Jigsaw Puzzle (SJP)—to evaluate their effectiveness in isolating distortion-related features. As shown in the left panel of Fig.~\ref{fig:aba_gms}, both Resize and GMS largely preserve the recognizable global structure of the scene (e.g., a woman or a tower). In contrast, SJP breaks the semantic layout into a stochastic arrangement of local patches, effectively disrupting high-level contextual cues. This scrambling mechanism prevents the backbone from relying on object recognition or scene semantics, thereby encouraging the model to concentrate on low-level textural patterns. Despite removing global structure, SJP maintains local degradation characteristics within each patch. Blurriness, noise, and compression artifacts remain clearly observable at the patch level. This behavior is quantitatively supported by the Sensitivity Ratio analysis shown on the right: SJP achieves a substantially higher distortion-to-semantic sensitivity ratio (1.10) compared to Resize and GMS (0.92–0.95), indicating stronger emphasis on degradation features. The raw CLIP response further demonstrates that SJP suppresses semantic similarity (blue bars) while preserving or enhancing distortion-related responses (orange bars), resulting in a cleaner distortion-aware embedding.

These results confirm that SJP functions as an effective semantic filter, enabling the Hierarchical Memory Consensus to retrieve neighbors based primarily on perceptual quality rather than semantic content bias.

\section{Additional Experiments on Robustness}
\label{sec:app_rob}
\subsection{Experiments on Effectiveness Analysis}
To evaluate the practical utility of our training-free framework, we compare the inference time of HIFC against several representative IQA methods, including both training-based~\cite{DBCNN,hyperiqa,styleam} and training-free models~\cite{clipiqa}. The results, measured on a single NVIDIA RTX 4090 GPU, are summarized in Table~\ref{tab:inference_time} and ~\ref{tab:efficiency}. Despite incorporating a hierarchical memory retrieval path, HIFC maintains a high inference speed of 14 ms per image. This is significantly faster than complex training-based models like StyleAM (128.6 ms) and comparable to the lightweight CLIP-IQA (13 ms). Such efficiency demonstrates that our dual-path architecture, combining rapid intuition with structured memory retrieval, does not introduce significant computational overhead. While some training-based methods achieve similar speeds, their performance lags far behind our train-free results. HIFC achieves the best balance in the efficiency-accuracy manifold, proving that a well-designed training-free paradigm can outperform traditional supervised models in both robustness and practical latency.
\begin{table}[h]
\centering
\caption{Comparison of Inference/Training costs and performance.}
\label{tab:inference_time}
\resizebox{0.7\columnwidth}{!}{
\begin{tabular}{lcccc}
\toprule
Method &Train& Inference Time(ms) & SRCC & PLCC\\
\midrule
HyperIQA  &$\checkmark$   & $16 \pm 3$&0.490 &0.487\\
DBCNN      & $\checkmark$ & $25 \pm 3$ &0.572&0.589\\
CLIP-IQA    &$\times$& $13 \pm 2$ &0.612&0.594\\
StyleAM       &$\checkmark$ & $128.6 \pm 3$&0.584 &0.561\\
\midrule
\textbf{Ours} &$\times$& $14 \pm 4$ &\textbf{0.778}&\textbf{0.780}\\
\bottomrule
\end{tabular}
}
\end{table}

\begin{table}[h]
\centering
\caption{Efficiency and complexity analysis of our method.}
\label{tab:efficiency}
\resizebox{0.9\columnwidth}{!}{
\begin{tabular}{lcc}
\toprule
Metric & Value & Description \\
\midrule
Backbone & CLIP (ViT-B/32) & Frozen, no training required \\
Parameters & 88 MB & Same as CLIP \\
FLOPs & $\sim$32.1G $\times 2$ & Dual-branch (parallel) \\
GPU Memory & 1118 MB & During inference \\
\midrule
Gallery Size & 10,125 & KADID-10k \\
Gallery Memory & 39 MB & Stored features (linear in $N$) \\
Build Time & $\sim$3 min & 27 it/s (offline, linear in $N$) \\
\midrule
Inference Time & $\sim$14 ms & Comparable to CLIP \\
Retrieval Time & 0.407 ms & $<$3\% of total runtime \\
\midrule
Complexity (Retrieval) & $O(N \cdot D)$ & FAISS inner-product search \\
Scalability & Linear in $N$ & Negligible overhead in practice \\
\bottomrule
\end{tabular}
}
\end{table}

\subsection{Performance on the Authentic-to-Authentic Setting}
We further evaluate HIFC in a purely authentic Authentic-to-Authentic cross-domain setting, where the memory bank is built from one authentic dataset (e.g., LIVEC) and tested on another (e.g., KONIQ), without synthetic distortion labels or explicit distortion alignment. As shown in Table~\ref{tab:a2a_res} , even under this constraint, HIFC achieves a strong SRCC of 0.755, outperforming existing SOTA BIQA methods and demonstrating that hierarchical retrieval with fuzzy consensus can capture shared perceptual regularities across authentic domains. Although a gap remains compared to fully supervised models such as TReS~\cite{Tres} in certain scenarios, this is mainly due to their domain-specific optimization via back-propagation and the residual semantic–distortion entanglement that arises when no explicit distortion supervision is available. Overall, the results confirm the robustness and practicality of the proposed training-free paradigm in real-world environments.
\begin{table}[t]
\centering
\caption{Performance evaluation of HIFC in the Authentic-to-Authentic cross-domain setting. The results demonstrate that HIFC achieves competitive performance compared with existing training methods by achieving high SRCC without any domain-specific training or synthetic supervision.}
\label{tab:a2a_res}
\resizebox{0.7\columnwidth}{!}{
\begin{tabular}{l c cc cc}
\toprule
\multirow{2}{*}{Reference} &\multirow{2}{*}{Method} 
& \multicolumn{2}{c}{LIVEC} & \multicolumn{2}{c}{KONIQ}\\
\cmidrule(lr){3-4} \cmidrule(lr){5-6}
& & SRCC & PLCC & SRCC & PLCC \\
\midrule
NONE & CLIP-IQA & 0.612 & 0.594 &0.695 & 0.717\\
\midrule
\multirow{3}{*}{LIVEC}
&TReS & -& -& 0.733   & 0.739  \\
&DBCNN &- & -& 0.754  & 0.759  \\
&Ours &- &- &\textbf{0.755} & \textbf{0.765}\\
\midrule
\multirow{3}{*}{KONIQ}
&TReS &\textbf{0.768} & \textbf{0.775}&- &-  \\
&DBCNN &0.755  & 0.769   & -  &-  \\   
&Ours &0.739 &0.765 &- &- \\
\bottomrule
\end{tabular}
}
\end{table}

\subsection{Performance on In-domain Setting}
We further evaluate HIFC in an In-domain setting, where the reference bank and query images are sampled from the same dataset with varying ratios. As shown in Table~\ref{tab:in_domain}, HIFC demonstrates exceptional stability across different datasets (KADID, KONIQ, and SPAQ~\cite{SPAQ}). Specifically, we observe that for synthetic datasets like KADID, a higher proportion of reference images (e.g., 9:1 ratio) yields the best performance (0.841 SRCC), as it provides a denser sampling of specific distortion levels. Conversely, for authentic datasets like KONIQ and SPAQ, the performance remains high and even improves as the query set. This suggests that HIFC effectively leverages the internal consistency of authentic distributions even with a relatively sparse reference bank. These results confirm that while HIFC is designed for zero-shot generalization, it can also serve as a highly effective in-domain quality estimator, accurately capturing the perceptual manifold of a specific target domain.
\begin{table}[h]
\centering
\caption{Performance evaluation of HIFC in the In-domain setting across various sampling ratios.}
\label{tab:in_domain}
\resizebox{0.7\columnwidth}{!}{
\begin{tabular}{c cc cc cc}
\toprule
\multirow{2}{*}{Ratio} 
& \multicolumn{2}{c}{KADID} 
& \multicolumn{2}{c}{KONIQ} & \multicolumn{2}{c}{SPAQ}\\
\cmidrule(lr){2-3} \cmidrule(lr){4-5} \cmidrule(lr){6-7}
& SRCC & PLCC & SRCC & PLCC & SRCC & PLCC\\
\midrule
9:1 & \textbf{0.841} &\textbf{0.852} & 0.822  &0.836  &0.880  & 0.882 \\
8:2 & 0.831 & 0.836  & \textbf{0.826}  &\textbf{ 0.839 }&\textbf{0.882}  &\textbf{0.883}\\
5:5 & 0.809 & 0.813  &0.825  &0.829  &0.873 &0.873 \\
2:8 & 0.778 & 0.786  &0.809& 0.815  &0.860  &0.861\\
1:9 &0.743 &0.758 &0.801  &0.803  &0.847  &0.843\\
\bottomrule
\end{tabular}
}
\end{table}

\subsection{Analysis of the Accuracy of the Top-5 Distortion}
The Top-5 retrieved distortion types (Table~\ref{tab:top5_dis}) provide strong evidence for the perceptual reliability of the Distortion Alignment (DA) module across diverse authentic datasets. For LIVEC and KONIQ, which contain images captured by various mobile devices, the frequent appearance of JPEG 2000, Motion Blur, and Gaussian Blur among the top-ranked distortions aligns well with common artifacts in consumer photography, such as compression effects and camera shake. In the BID dataset, which primarily focuses on blur, the DA module accurately identifies Motion Blur and Lens Blur as dominant factors. For PIPAL, which contains complex GAN-based and traditional distortions, the frequent retrieval of Motion Blur and JPEG 2000 indicates that the module effectively decomposes complex artifacts into perceptually similar fundamental degradation types. The substantial overlap of top-ranked distortion types across datasets (e.g., Motion Blur consistently appearing within the Top-2) confirms that the distorted-memory bank captures shared perceptual structures of authentic degradations. Even without ground-truth distortion labels for real images, the DA module successfully projects them onto a structured synthetic distortion manifold, ensuring that subsequent hierarchical retrieval is grounded in objective degradation characteristics rather than semantic bias.

\begin{table}[h]
\centering
\caption{Analysis of Top-5 retrieved distortion types across diverse datasets. The alignment results demonstrate that the DA module accurately maps complex unknown degradations to perceptually similar synthetic prototypes.}
\label{tab:top5_dis}
\resizebox{\columnwidth}{!}{
\begin{tabular}{lccccc}
\toprule
Dataset &Top-1&Top-2&Top-3&Top-4&Top-5\\
\midrule
LIVEC  &\textbf{JPEG 2000}&\textbf{Motion Blur}&JPEG&Darken&Pixelate\\
KONIQ  &\textbf{Motion Blur}&Gaussian Blur&\textbf{JPEG 2000}&Denoise&Color Diffusion\\
BID   &Denoise&\textbf{Motion Blur}&Gaussian Blur&Lens Blur&\textbf{JPEG 2000}\\
PIPAL&\textbf{Motion Blur}&\textbf{JPEG 2000}&JPEG&Gaussian Blur&Contrast\\
\bottomrule
\end{tabular}
}
\end{table}

\section{Additional Cross-Domain and Few-Shot Generalization Experiments}
\label{sec:app_cross}
\label{sec:additional_cross_domain}
We further evaluate the generalization capability of HIFC in several challenging cross-domain scenarios that go beyond the settings covered in the main paper. In the first experiment, we assess performance on datasets exhibiting diverse and non-traditional distortion characteristics, including AI-generated content, underwater and high-resolution datasets. As summarized in Table~\ref{tab:cross_domain_aigc}, HIFC, using only KADID-10k as the reference gallery, is compared against several training-based methods that were fully supervised on KADID-10k. Despite being entirely training-free, HIFC achieves the best PLCC on AGIQA-3K, AGIQA-20K and SPAQ, and competitive performance on SAUD. These results demonstrate the strong zero-shot transferability of the quality-trace prototypes extracted by the hierarchical memory consensus, even when the target domain contains distortions significantly different from those in the reference set.

A second experiment extends the evaluation to aesthetic quality assessment, where the notion of quality is heavily influenced by subjective and cultural factors. Table~\ref{tab:cross_domain_aesthetic} reports the SRCC and PLCC on the AVA and PARA datasets, again using the same KADID-10k reference bank. HIFC surpasses dedicated aesthetic assessment methods such as NIMA and DeQA-Score on the PARA benchmark, while maintaining reasonable performance on AVA. The pronounced advantage on PARA suggests that the fuzzy consensus mechanism can, to a certain extent, capture the low-level perceptual regularities that generalize across both quality and aesthetic domains, despite the absence of artistic priors in the reference bank.

Finally, we examine the few-shot adaptability of HIFC by varying the source dataset used to build the reference bank, and compare it with the training-based PromptIQA method~\cite{chen2024promptiqa} evaluated on KonIQ-10k. As shown in Table~\ref{tab:koniq_comparison}, when LIVE, CSIQ, or KADID-10k serve as the sole reference source, HIFC consistently outperforms PromptIQA, which requires end-to-end training on the corresponding source domain. These results highlight that the proposed framework can function as a robust few-shot quality estimator: by simply substituting the reference gallery with a small set of annotated examples from a related domain, HIFC achieves strong alignment with target perception without any gradient updates, offering a practical plug-and-play solution for real-world IQA deployment.

\begin{table}[h]
\centering
\caption{Cross-domain generalization performance (PLCC). Other models are trained on KADID and evaluated on underwater/AIGC/high-resolution datasets.}
\label{tab:cross_domain_aigc}
\resizebox{0.75\columnwidth}{!}{
\begin{tabular}{lcccc}
\toprule
Method & AGIQA-3K & AGIQA-20K & SAUD  & SPAQ \\
\midrule
BRISQUE   & 0.178 & 0.118 & 0.216 &  0.314 \\
DBCNN     & 0.324 & 0.284 & 0.418 &  0.559 \\
HyperIQA   &0.748  & 0.516  &0.497  & 0.688  \\
MetaIQA     &0.468 & 0.348 & 0.586 &  0.646 \\
UCDA      & 0.681 &0.255  &0.627  &  0.699 \\
DGQA  & 0.577 & 0.498  & 0.651 & 0.705  \\
\midrule
\textbf{Ours(Ref on KADID-10k)} & \textbf{0.759} & \textbf{0.671} &\textbf{0.539}  & \textbf{0.830} \\
\bottomrule
\end{tabular}
}
\end{table}

\begin{table}[t]
\centering
\caption{Cross-domain generalization performance (SRCC/PLCC). Models are trained on KADID and evaluated on aesthetic datasets.}
\label{tab:cross_domain_aesthetic}
\resizebox{0.65\columnwidth}{!}{
\begin{tabular}{lcc}
\toprule
Method & AVA & PARA \\
\midrule
NIMA   &\textbf{0.612}/\textbf{0.636} & 0.571/0.587  \\
BRISQUE      &0.031/0.036  &0.078/0.129 \\
Q-Align  &0.521/0.541  &0.626/0.618  \\
DeQA-Score      &0.572/0.589  &0.680/0.703 \\
DGQA  & -/- &0.322/0.351  \\
CLIP-IQA  & 0.320/0.317 &0.535/0.566  \\
\midrule
\textbf{Ours(Ref on KADID-10k)} &0.457/0.468 &\textbf{0.684}/\textbf{0.713} \\
\bottomrule
\end{tabular}
}
\end{table}

\begin{table}[htbp]
  \centering
  \caption{Performance comparison on the KonIQ-10k dataset when training/referencing on three different datasets from synthetic to real. The best results are highlighted in \textbf{bold}.}
  \label{tab:koniq_comparison}
  \begin{tabular}{llccc}
    \toprule
    \multirow{2}{*}{Train/Reference} & \multirow{2}{*}{Method} &\multirow{2}{*}{Train}& \multicolumn{2}{c}{KonIQ-10k} \\
    \cmidrule(lr){4-5}
    & & & SRCC & PLCC \\
    \midrule
    \multirow{2}{*}{LIVE} & PromptIQA & \checkmark  & 0.660 & 0.698 \\
    & Ours & $\times$ & \textbf{0.696} & \textbf{0.707} \\
    \midrule
    \multirow{2}{*}{CSIQ} & PromptIQA& \checkmark & 0.619 & 0.629 \\
    & Ours& $\times$ & \textbf{0.693} & \textbf{0.714} \\
    \midrule
    \multirow{2}{*}{KADID-10k} & PromptIQA& \checkmark & 0.577 & 0.595 \\
    & Ours& $\times$ & \textbf{0.765} & \textbf{0.759} \\
    
    \bottomrule
  \end{tabular}
\end{table}
\subsection{Why Stacking for the Semantic Branch?}
\label{subsec:stacking}

The semantic branch is designed to retain content and scene context, complementing the distortion branch, which deliberately disrupts global spatial topology. Direct resizing can alter aspect ratios and attenuate local details, while center cropping can omit peripheral objects and contextual information. Sliding-window stacking aggregates content sampled across the image and reduces reliance on a single crop.

The controlled comparison in Table~\ref{tab:semantic-preprocessing} fixes the backbone, gallery, distortion branch, retrieval, and fusion. Stacking achieves the highest prediction correlations among the evaluated alternatives. Its measured preprocessing latency is approximately 1.09 ms for $512\times512$ inputs on an RTX 4090, the lowest among the tested configurations. This timing concerns preprocessing only, rather than total inference latency.

\begin{table}[htbp]
\centering
\caption{Controlled semantic preprocessing comparison. Prediction results are SRCC/PLCC. Latency measures sampling/preprocessing for $512\times512$ images on an RTX 4090. Multiple patches are batched for inference. Stacking is implemented with \texttt{torch.unfold}; the other alternatives use transform-library implementations.}
\label{tab:semantic-preprocessing}
\small
\begin{tabular}{@{}lccc@{}}
\toprule
Semantic input & LIVEC & KonIQ-10k & Preprocessing (ms) \\
\midrule
Direct resize & 0.692/0.720 & 0.682/0.687 & 3.439 \\
Letterbox resize & 0.712/0.733 & 0.671/0.675 & 3.483 \\
Center crop & 0.728/0.745 & 0.707/0.717 & 3.488 \\
Five random crops, averaged & 0.726/0.742 & 0.716/0.719 & 2.026 \\
Five-crop average & 0.773/0.778 & 0.751/0.752 & 1.786 \\
Proposed stacking & 0.778/0.780 & 0.765/0.759 & 1.086 \\
\bottomrule
\end{tabular}
\end{table}
\section{Analysis on Failure Case}
\label{sec:fail}
\begin{figure}[h]
    \centering
    \includegraphics[width=0.8\linewidth]{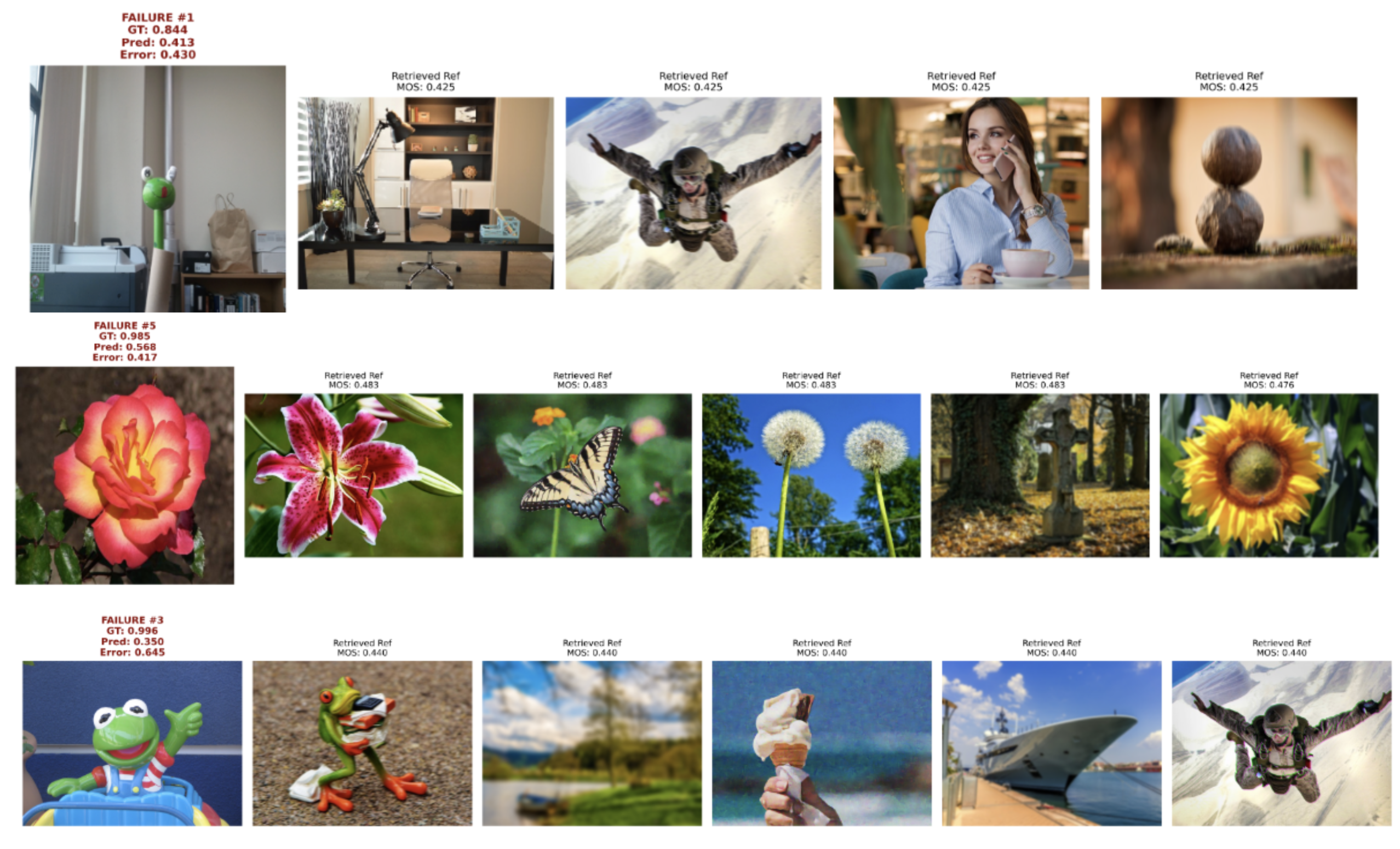}
    \caption{Visual analysis of failure cases. The HIFC framework occasionally produces biased estimations when encountering images with dominant semantic aesthetics (e.g., artistic bokeh) or extreme lighting conditions, which the distortion-related features fail to fully decouple from physical degradations.}
    \label{fig:fail}
\end{figure}

While the HIFC framework demonstrates strong robustness across diverse benchmarks, its performance depends critically on the reliability of the retrieved memory manifold. In certain edge cases, the retrieval process may be influenced by dominant visual attributes that are weakly correlated with actual degradation. Fig.~\ref{fig:fail} presents three representative failure examples.

\textbf{Case 1: Global Color and Illumination Bias (Top Row).}  
In the first example (GT: 0.844, Pred: 0.413), the query is a high-quality indoor image. However, the retrieved references exhibit highly similar global color tones and illumination patterns, such as bright white backgrounds and strong contrast lighting. When global color statistics become overly dominant, they can overshadow subtle textural degradation cues. As a result, the model retrieves a cluster of images with comparable lighting conditions but substantially lower perceptual quality (MOS $\approx$ 0.425), leading to an underestimated consensus score.

\textbf{Case 2 \& 3: Semantic and Compositional Override (Middle and Bottom Rows).}  
The second and third examples illustrate the issue of \emph{semantic override}. In the middle row (a macro photograph of a rose), the retrieval favors images with similar composition, such as centrally framed subjects with shallow depth-of-field backgrounds. In the bottom row, a toy frog query retrieves another toy frog as its nearest neighbor. Despite the semantic-distortion decoupling introduced by the DA and SDR modules, extreme semantic or compositional similarities can still dominate the feature distance metric. Consequently, the memory bank prioritizes object category or photographic style over degradation characteristics, resulting in a biased quality centroid.

\textbf{Insights.}  
These cases highlight an important limitation: although the training-free, zero-shot paradigm generally aligns well with human perception, it may struggle when highly salient semantics or extreme color distributions dominate the embedding space, especially if such patterns are sparsely represented in the reference bank. Mitigating this residual semantic entanglement in a fully unsupervised setting remains an open research challenge.

\section{Limitations and Future Work}
\label{sec:lim}

While HIFC achieves competitive train-free cross-domain IQA
performance, several limitations remain. As revealed in the
failure analysis, extreme semantic structures or dominant color
and illumination patterns can occasionally bias retrieval.
Scrambled jigsaw preprocessing disrupts global spatial topology
but does not fully eliminate these cues. As shown in
Table~\ref{tab:color-preprocessing}, grayscale and luminance-only
preprocessing modestly improve SRCC on the color/illumination
subset of TID2013, but reduce overall prediction performance.
This suggests that suppressing chromatic information can
alleviate certain biases while also discarding genuine quality
cues, such as color casts, chromatic noise, and saturation
artifacts. We therefore retain RGB preprocessing and identify
adaptive RGB--luminance representations, together with more
principled content--distortion decoupling, as promising directions
for strengthening distortion-focused retrieval.

\begin{table}[t]
    \centering
    \caption{Effect of distortion-branch preprocessing on TID2013.
    SJP denotes scrambled jigsaw puzzle preprocessing.}
    \label{tab:color-preprocessing}
    \resizebox{\linewidth}{!}{
    \begin{tabular}{lccc}
        \toprule
        Distortion input
        & Overall SRCC
        & Overall PLCC
        & Color/illumination SRCC \\
        \midrule
        RGB SJP
        & 0.6499 & 0.6569 & 0.3759 \\
        Grayscale SJP
        & 0.6405 & 0.6442 & 0.3932 \\
        Luminance-only SJP
        & 0.6432 & 0.6483 & 0.3997 \\
        \bottomrule
    \end{tabular}
    }
\end{table}

Although the main experiments use a unified hyperparameter
configuration across target domains, parameters such as
$N_{\text{dist}}$ and $k$ are still manually specified.
Developing a self-adaptive mechanism that responds to the
statistical properties of the distortion manifold could further
enhance the plug-and-play capability of the framework and reduce
human intervention.

We also plan to investigate advanced representation learning
strategies, such as contrastive learning with quality-preserving
augmentations, to further refine the embedding space.
Such approaches could improve the separation between semantic
content and degradation characteristics, strengthening the
alignment between memory retrieval and perceptual quality cues.

\newpage
\section{Pytorch-like Pseudocode for HIFC}
\begin{algorithm}[h]
\caption{HIFC Inference}
\label{alg:hifc_pytorch}
\KwIn{Query image $I_q$, reference bank $\mathcal{G}$, CLIP model $\Phi$}
\KwOut{Final quality score $\hat{Q}$}

\textbf{\# Hyperparameters} $N_{\text{dist}},\; K,\; \alpha$
\vspace{0.3em}

\textbf{def HIFC\_Inference($I_q$, $\mathcal{G}$, $\Phi$):}\\
\hspace{1em} \texttt{\# 1. Feature Extraction \& Decoupling} \\
\hspace{1em} $q\_sem \;=\; \Phi.\texttt{encode\_image}(\texttt{extract\_patches}(I_q)).\texttt{mean}(0)$ \\
\hspace{1em} $q\_dis \;=\; \Phi.\texttt{encode\_image}(\texttt{SJP\_transform}(I_q))$

\vspace{0.3em}
\hspace{1em} \texttt{\# 2. Level-1: Distortion Alignment (DA)} \\
\hspace{1em} $dist\_centroids \;=\; \mathcal{G}.\texttt{groupby}('dist\_type').f\_dis.\texttt{mean}()$ \\
\hspace{1em} $sim\_dist \;=\; \texttt{cosine\_similarity}(q\_dis, dist\_centroids)$ \\
\hspace{1em} $top\_dist \;=\; sim\_dist.\texttt{topk}(N_{\text{dist}}).\texttt{indices}$ \\
\hspace{1em} $\mathcal{G}^* \;=\; \mathcal{G}[\mathcal{G}.\texttt{dist\_type} \in top\_dist]$

\vspace{0.3em}
\hspace{1em} \texttt{\# 3. Level-2: Semantic-Guided Retrieval (SGR)} \\
\hspace{1em} $candidates \;=\; \texttt{search\_index}(q\_sem, \mathcal{G}^*.f\_sem, top\_n=1000)$

\vspace{0.3em}
\hspace{1em} \texttt{\# 4. Level-3: Memory Quality-Centroid Consensus (QCC)} \\
\hspace{1em} $local\_sim \;=\; \texttt{cosine\_similarity}(q\_dis, \mathcal{G}^*[candidates].f\_dis)$ \\
\hspace{1em} $neighbors \;=\; local\_sim.\texttt{topk}(K).\texttt{indices}$ \\
\hspace{1em} $score\_mem \;=\; \mathcal{G}^*[neighbors].MOS.\texttt{mean}()$

\vspace{0.3em}
\hspace{1em} \texttt{\# 5. Zero-Shot Intuition (ZSI)} \\
\hspace{1em} $score\_int \;=\; \Phi.\texttt{compute\_text\_alignment}(q\_sem,\;["good","bad"])$

\vspace{0.3em}
\hspace{1em} \texttt{\# 6. Final Fusion} \\
\hspace{1em} $\hat{Q} \;=\; \alpha \cdot score\_mem \;+\; (1-\alpha)\cdot score\_int$

\vspace{0.3em}
\hspace{1em} \textbf{return} $\hat{Q}$

\end{algorithm}

\section{Comparison with Additional VLM-based Methods.}
We extend the comparison to recent VLM-based IQA methods,
including CLIP-IQA, MACLIP, LIQE, Q-Align,
DeQA-Score, Qwen2.5-VL-7B, and Q-insight.
As shown in Table~\ref{tab:vlm-comparison}, HIFC-IQA
achieves the highest SRCC of 0.778 on LIVEC and maintains
competitive performance on KonIQ-10k and BID.
Q-insight achieves the best correlations on KonIQ-10k and BID,
while HIFC-IQA offers a lightweight alternative without
IQA-specific parameter training. We further evaluate these methods on four additional datasets covering AI-generated image quality, aesthetic assessment,
and image restoration. HIFC-IQA achieves the highest SRCC on AGIQA-3K (0.759),
PARA (0.684), and PIPAL (0.507), while DeQA-Score achieves the best result on AVA. These results demonstrate the generalization capability of HIFC-IQA across diverse perceptual quality domains, highlighting the effectiveness of combining reference-based quality consensus with zero-shot perceptual priors.

\begin{table*}[b]
\centering
\caption{Comparison with additional VLM-based IQA methods across
diverse perceptual quality datasets. KADID-10k serves as the
IQA training source or reference gallery.
Train indicates whether IQA-specific parameter training is required.
SRCC and PLCC are reported for LIVEC, KonIQ-10k, and BID,
while SRCC is reported for the remaining datasets.}
\label{tab:vlm-comparison}
\resizebox{0.85\textwidth}{!}{
\begin{tabular}{l c cc cc cc cccc}
\toprule
\multirow{2}{*}{Methods}
& \multirow{2}{*}{Train}
& \multicolumn{2}{c}{LIVEC}
& \multicolumn{2}{c}{KonIQ-10k}
& \multicolumn{2}{c}{BID}
& \multicolumn{4}{c}{Additional Domains (SRCC)} \\
\cmidrule(lr){3-4}
\cmidrule(lr){5-6}
\cmidrule(lr){7-8}
\cmidrule(lr){9-12}
& & SRCC & PLCC & SRCC & PLCC & SRCC & PLCC
& AGIQA-3K & AVA & PARA & PIPAL \\
\midrule

% CLIP-IQA~\cite{clipiqa}
% & No
% & 0.612 & 0.594
% & 0.695 & 0.617
% & 0.688 & 0.711
% & 0.696 & 0.320 & 0.535 & 0.427 \\

MACLIP
& No
& 0.679 & 0.703
& 0.738 & 0.738
& 0.592 & 0.547
& 0.753 & 0.298 & 0.504 & 0.379 \\

LIQE
& Yes
& 0.677 & 0.680
& 0.719 & 0.726
& 0.684 & 0.652
& 0.653 & 0.429 & 0.483 & 0.480 \\

Q-Align~\cite{qalign}
& Yes
& 0.702 & 0.744
& 0.668 & 0.665
& 0.576 & 0.651
& 0.733 & 0.521 & 0.626 & 0.403 \\

DeQA-Score
& Yes
& 0.743 & 0.775
& 0.677 & 0.703
& 0.702 & 0.743
& 0.745 & \textbf{0.572} & 0.680 & 0.472 \\

Qwen2.5-VL-7B
& Yes
& 0.761 & 0.760
& 0.754 & 0.810
& 0.711 & 0.725
& -- & -- & -- & -- \\

Q-insight
& Yes
& 0.761 & \textbf{0.795}
& \textbf{0.806} & \textbf{0.829}
& \textbf{0.784} & \textbf{0.796}
& -- & -- & -- & -- \\

\midrule
HIFC-IQA
& \textbf{No}
& \textbf{0.778} & 0.780
& 0.765 & 0.759
& 0.744 & 0.749
& \textbf{0.759} & 0.457
& \textbf{0.684} & \textbf{0.507} \\

\bottomrule
\end{tabular}
}
\end{table*}

\section{Relation to Retrieval-Based IQA}
\label{sec:retrieval-related}

Retrieval-based IQA encompasses several distinct paradigms. IQARAG~\cite{fu2026iqarag} retrieves reference images and their MOSs as visual anchors for LMM score generation. QualiRAG~\cite{cao2026qualirag} retrieves dynamically generated auxiliary knowledge for visual quality understanding. Wang et al.~\cite{wang2023regressionfree} construct hierarchical retrieval through separate semantic and distortion classification modules and aggregate retrieved subjective scores. Table~\ref{tab:retrieval-families} summarizes these differences.

HIFC-IQA constructs asymmetric semantic and distortion representations using a frozen CLIP backbone. It combines target-to-source distortion alignment, semantic candidate filtering, distortion-aware reranking, and adaptive fuzzy quality consensus (FQC). The memory score is then fused with an independent zero-shot intuition score. Its contribution therefore extends beyond introducing a memory bank or averaging the labels of nearest neighbors.

\begin{table}[htbp]
\centering
\caption{Comparison of retrieval-based IQA paradigms. MOS denotes mean opinion score; LMM denotes large multimodal model.}
\label{tab:retrieval-families}
\small
\begin{tabular}{@{}p{0.17\linewidth}p{0.25\linewidth}p{0.22\linewidth}p{0.28\linewidth}@{}}
\toprule
Method & Retrieved information & Prediction & Distinguishing mechanism \\
\midrule
IQARAG~\cite{fu2026iqarag} & Reference images and MOSs & LMM score generation & Visual in-context quality anchors \\
QualiRAG~\cite{cao2026qualirag} & Metadata, localization, and quality descriptions & Evidence-grounded LMM response & Retrieval of auxiliary quality knowledge \\
Wang et al.~\cite{wang2023regressionfree} & Subjectively scored images & Hierarchical kNN and MOS aggregation & Separate semantic and distortion classification modules \\
HIFC-IQA & Synthetic images and MOSs & FQC and zero-shot intuition & Frozen dual-space alignment and adaptive fuzzy consensus \\
\bottomrule
\end{tabular}
\end{table}

\clearpage  % TODO FINAL: This \clearpage needs to be removed from both review and camera-ready versions.


%%%%%%%%%%%%%%%%%%%%%%%%%%%%%%%%%%%%%%%%%%%%%%%%%%%%%%%%%%%%

\newpage
\section*{NeurIPS Paper Checklist}

\begin{enumerate}

\item {\bf Claims}
    \item[] Question: Do the main claims made in the abstract and introduction accurately reflect the paper's contributions and scope?
    \item[] Answer: \answerYes{} % Replace by \answerYes{}, \answerNo{}, or \answerNA{}.
    \item[] Justification: The abstract and introduction present the contributions.

\item {\bf Limitations}
    \item[] Question: Does the paper discuss the limitations of the work performed by the authors?
    \item[] Answer: \answerYes{} % Replace by \answerYes{}, \answerNo{}, or \answerNA{}.
    \item[] Justification: In Appendix Sec.H and Sec.I.

\item {\bf Theory assumptions and proofs}
    \item[] Question: For each theoretical result, does the paper provide the full set of assumptions and a complete (and correct) proof?
    \item[] Answer: \answerYes{} % Replace by \answerYes{}, \answerNo{}, or \answerNA{}.
    \item[] Justification: In Appendix Sec.C. Non-rigorous mathematical proof, simple description and motivation rationality of the conclusion through the phenomenon found in other articles.

\item {\bf Experimental result reproducibility}
    \item[] Question: Does the paper fully disclose all the information needed to reproduce the main experimental results of the paper to the extent that it affects the main claims and/or conclusions of the paper (regardless of whether the code and data are provided or not)?
    \item[] Answer: \answerYes{} % Replace by \answerYes{}, \answerNo{}, or \answerNA{}.
    \item[] Justification:  In Section 4, we detail the experimental setup, along with pytorch-like pseudocode.

\item {\bf Open access to data and code}
    \item[] Question: Does the paper provide open access to the data and code, with sufficient instructions to faithfully reproduce the main experimental results, as described in supplemental material?
    \item[] Answer: \answerYes{} % Replace by \answerYes{}, \answerNo{}, or \answerNA{}.
    \item[] Justification: The code will be available upon acceptance.

\item {\bf Experimental setting/details}
    \item[] Question: Does the paper specify all the training and test details (e.g., data splits, hyperparameters, how they were chosen, type of optimizer) necessary to understand the results?
    \item[] Answer: \answerYes{} % Replace by \answerYes{}, \answerNo{}, or \answerNA{}.
    \item[] Justification:  In Section 4 and Appendix B

\item {\bf Experiment statistical significance}
    \item[] Question: Does the paper report error bars suitably and correctly defined or other appropriate information about the statistical significance of the experiments?
    \item[] Answer: \answerYes{} % Replace by \answerYes{}, \answerNo{}, or \answerNA{}.
    \item[] Justification: In Section 4

\item {\bf Experiments compute resources}
    \item[] Question: For each experiment, does the paper provide sufficient information on the computer resources (type of compute workers, memory, time of execution) needed to reproduce the experiments?
    \item[] Answer: \answerYes{} % Replace by \answerYes{}, \answerNo{}, or \answerNA{}.
    \item[] Justification:  In Section 4 and Appendix B.
    
\item {\bf Code of ethics}
    \item[] Question: Does the research conducted in the paper conform, in every respect, with the NeurIPS Code of Ethics \url{https://neurips.cc/public/EthicsGuidelines}?
    \item[] Answer: \answerYes{} % Replace by \answerYes{}, \answerNo{}, or \answerNA{}.
    \item[] Justification: The research conforms to the NeurIPS Code of Ethics

\item {\bf Broader impacts}
    \item[] Question: Does the paper discuss both potential positive societal impacts and negative societal impacts of the work performed?
    \item[] Answer: \answerNA{} % Replace by \answerYes{}, \answerNo{}, or \answerNA{}.
    \item[] Justification:  There is no societal impact of the work performed.
    
\item {\bf Safeguards}
    \item[] Question: Does the paper describe safeguards that have been put in place for responsible release of data or models that have a high risk for misuse (e.g., pre-trained language models, image generators, or scraped datasets)?
    \item[] Answer: \answerNA{} % Replace by \answerYes{}, \answerNo{}, or \answerNA{}.
    \item[] Justification: The paper poses no such risks

\item {\bf Licenses for existing assets}
    \item[] Question: Are the creators or original owners of assets (e.g., code, data, models), used in the paper, properly credited and are the license and terms of use explicitly mentioned and properly respected?
    \item[] Answer: \answerYes{} % Replace by \answerYes{}, \answerNo{}, or \answerNA{}.
    \item[] Justification: Properly cited and stated

\item {\bf New assets}
    \item[] Question: Are new assets introduced in the paper well documented and is the documentation provided alongside the assets?
    \item[] Answer: \answerYes{} % Replace by \answerYes{}, \answerNo{}, or \answerNA{}.
    \item[] Justification: In Section 4 and Appendix B

\item {\bf Crowdsourcing and research with human subjects}
    \item[] Question: For crowdsourcing experiments and research with human subjects, does the paper include the full text of instructions given to participants and screenshots, if applicable, as well as details about compensation (if any)? 
    \item[] Answer: \answerNA{} % Replace by \answerYes{}, \answerNo{}, or \answerNA{}.
    \item[] Justification: Not related

\item {\bf Institutional review board (IRB) approvals or equivalent for research with human subjects}
    \item[] Question: Does the paper describe potential risks incurred by study participants, whether such risks were disclosed to the subjects, and whether Institutional Review Board (IRB) approvals (or an equivalent approval/review based on the requirements of your country or institution) were obtained?
    \item[] Answer: \answerNA{} % Replace by \answerYes{}, \answerNo{}, or \answerNA{}.
    \item[] Justification: Not related

\item {\bf Declaration of LLM usage}
    \item[] Question: Does the paper describe the usage of LLMs if it is an important, original, or non-standard component of the core methods in this research? Note that if the LLM is used only for writing, editing, or formatting purposes and does \emph{not} impact the core methodology, scientific rigor, or originality of the research, declaration is not required.
    %this research? 
    \item[] Answer: \answerNA{} % Replace by \answerYes{}, \answerNo{}, or \answerNA{}.
    \item[] Justification: Not related

\end{enumerate}

\end{document}